\documentclass[
  11pt,
  a4paper,
  onecolumn
]{article}

\usepackage[english]{babel}
\usepackage[T1]{fontenc}
\usepackage{times}
\usepackage{amsmath}
\usepackage{amsfonts}
\usepackage{lipsum}
\usepackage{subcaption}
\usepackage{booktabs}
\usepackage{tabularx}
\usepackage{tikz}
\usetikzlibrary{positioning}
\usepackage{longtable}
\usepackage{array}
\usepackage{geometry}
\usepackage{adjustbox}
\usepackage{hyperref}
\usepackage{cleveref}
\usepackage{placeins}
\title{Accelerated surrogate dynamics for dynamical, stochastic system evolution}

\author{
  Marco Jochum$^{1,2}$
  \and
  Ioannis Kouroudis$^{1,2}$\thanks{ioannis.kouroudis@tum.de}
  \and
  Gohar Ali Siddiqui$^{1,2}$
  \and
  Taher Amine Hamzaoui$^{1}$
  \and
  Manuel ~Gö{\ss}wein$^{1}$
  \and
  Prof.\ Dr.\ rer.\ nat.\ Alessio Gagliardi$^{1}$%
  \thanks{alessio.gagliardi@tum.de}
}

\date{ }

\begin{document}

% \textcolor{red}{Gohar}
% \textcolor{blue}{Ioannis}
% \textcolor{yellow}{Alessio}

\maketitle

\begin{center}
\small
$^{1}$ Chair of Simulation of Nanosystems for Energy Conversion,
Department of Electrical Engineering, TUM School of Computation,
Information and Technology, Atomistic Modeling Center (AMC),
Munich Data Science Institute (MDSI),
Technical University of Munich,
Hans-Piloty-Straße 1, 85748 Garching, Germany

\vspace{0.5em}

$^{2}$ These authors contributed equally
\end{center}

\begin{abstract}
  Dynamic simulations are an entrenched way of gaining insight into the evolution of system dynamics. Their computational cost however is often prohibitively high, especially in cases of stochastic frameworks. Machine learning algorithms are especially suited as simulation surrogates. Nevertheless, they face some very distinct limitations. Firstly, the sheer dimensionality of these systems, however, precludes the use of traditional time series  models who struggle with high dimensional feature spaces. Additionally, traditional time series focus exclusively on either long or short range effects, causing local or global drift given enough time. In this paper, we propose a framework that addresses those limitations. Our framework combines a Variational Autoencoder, with a convolutional or graph basis that reduces the dimensionality of the system. This latent vector is propagated in time using a Temporal Fusion Transformer model, which includes both long range and short range effect encoding, as well as static covariate support. We test our framework on three distinct cases, to prove its robustness and in all three we have achieved practically identical to the simulation results at a fraction of the time. Further, our framework is flexible enough to be adapted to any new system and provides an inbuilt uncertainty quantification for targeted experiment design.
\end{abstract}

\section{Introduction}
Phenomena investigation is traditionally undertaken by laborious and meticulously designed experiments. Frequently however, these in themselves are not sufficiently able to isolate the underlying mechanisms that cause the observed dynamics.  
The systematic understanding of natural phenomena therefore is more deeply achieved through time-consuming physical simulations such as Kinetic Monte Carlo (KMC) \cite{katzenmeier2022modeling} and molecular dynamics (MD) \cite{hollingsworth2018molecular}. One significant drawback however is that to understand the physical system fully, a high computational and time load is necessary. Traditionally, this issue is mitigated in two ways. The first focuses on developing simulation tools that are more highly parallelizable, thus increasing the hardware utilization and decreasing the simulation time \cite{parallel1, parallel2, parallel3}. The second is more mathematical and attempts to approximate simulation elements with coarser and less expensive ones, with a controllable loss of accuracy. This can happen both on the level of domain discretization \cite{refinement1, refinement2} or on the number of necessary simulations \cite{coarse_graining_1, coarse_graining_3, coarse_graining_2}. Nevertheless, a highly promising and emergent third way is the use of Machine Learning to either accelerate or outright substitute the simulation results \cite{mayr2022machine}. The simplest approach is to decrease the number of simulation runs necessary. This can for example take the form of efficiently determining the system parametrization that matches the experimental results\cite{kouroudis2023utilizing, vernickel2020machine}, or accelerating the trial and error process of optimal system configurations \cite{kouroudis2025augur,todorovic2019bayesian}. Further, it can provide a coarse-grained representation of high informational content such as in the case of \cite{siddiqui2023application, wang2019machine}. Lastly, it can be used in a hybrid way to approximate a computationally expensive simulation step, while keeping the remaining simulation physics driven \cite{kochkov2021machine,friederich2021machine}. Nevertheless, the creation of a complete simulation surrogate has proven a much harder proposition. A good overview of the proposed solutions can be found in \cite{noe2020machine}. Long Short Term Memory (LSTM) networks were used for dynamics surrogation such as Molecular Dynamics \cite{LSMT1, LSTM2} and Kinetic Monte Carlo \cite{LSTMKMC1}. Intrinsically however, these architectures focus on short term dynamical trends, neglecting the long term effects. On the obverse, transformers are an algorithm allows the capture for long term dynamics, at the expense of short term information. This has been successfully used in dynamics simulations \cite{transformers_md_1, transformers_md_2}, albeit with the aforementioned limitations. Further, the above mentioned algorithms scale badly, both in accuracy and in computational cost, with increasing dimensionality inputs. 
Therefore, both LSTMs and transformers suffer significantly when deployed in complex applications where the geometry is high dimensional and the effects both long and short term. 
Additionally, these algorithms are opaque, providing predictions without any explainability behind them. To address all these issues,  we developed a pipeline that simultaneously trains a dimensionality reduction part with the encoding branch of a  Variational AutoEncoder (VAE), a time propagator in the form of Temporal Fusion Transformers and a dimensionality reconstruction through the decoding branch of the VAE. The VAE reduces the dimensionality to a manageable degree, the temporal fusion transformer combines LSTMs and transformers for capturing every dynamical scale and the decoder projects the results back to the original dimensionality.  We show that this algorithm fusion is able to efficiently and accurately propagate trajectories of diverse simulation methods and cases. Additionally, through the unique variable selection capabilities of the propagator, we are able to assign decipher the importance of static parameters to the phenomenon evolution thus adding a degree of explainability to the pipeline.  We deploy our pipeline in 3 diverse dynamical cases to showcase its versatility and robustness under highly different applications. We show that our pipeline can be successfully deployed in both stochastic and deterministic dynamical systems , presented both as space configurations and as molecular graphs. 

The first and simplest is the case of spin exchange. This simulation method has been given comparatively little importance in the literature \cite{liu2022advancing} and is meant to showcase the first principles of our algorithm.  It nevertheless has some highly interesting applications such as water desalination \cite{desalination} and hydrogen separation \cite{fuel_cells}.

% Nevertheless, there is a level of complexity that provides some interesting techniques. Namely, unlike the remaining cases, multiple different ion configurations can lead to valid time steps. This creates an issue as the algorithm is called to create a prediction for which multiple outputs are valid. Averaging the pixel map is not possible (as opposed to the kinetic monte carlo case) as the different ion would result to noise average, even in structured profiles. To this end, we predicted the system energy and used it to inform a ResNet architecture which proved crucial to a consistent and robust evolution prediction.
% # clustering of organics and magnetization

The second is focused around a kinetic Monte Carlo (kMC) model. This is commonly used to perform ensemble based simulations particularly for solid state electrolytes \cite{SSE} and organic semi conductors \cite{SEMICOND}. 
Few methods have been developed to address the computational limitations of kMC with data-driven approaches and none to our knowledge reconstructs the full stochastic profile of the simulation which our framework provides. 

The last case is an atomistic simulation using molecular dynamics (MD). MD has been proven over the years to be a central tool in the study of dynamic behaviour of matter. 
% Thus, a library of methods has been developed to optimize and accelerate MD \cite{lu2021,jones2022}. The bulk of computational effort in MD is spent in the calculation of energy and forces. These can come from a quantum mechanical simulation e.g. DFT, an parameterized empirical model or more recently, a machine learning model. In any case, large spatial and temporal simulations require prohibitively large computational resource. 
A major reason for long simulations is sampling rate parity between the different events during the dynamics \cite{iyengar2024}. This is closely related to the well-known "Sampling Problem" in statistics. 
% In this case, much of the computational time is spent in well-sampled region of the configuration space until a \textit{rare} event enables a transition. Although important to get the correct population of states, the well-sampled states do not explore any new configuration space and thus can be replaced by a dynamic surrogate model. 
% To demonstrate our algorithm, we take the well known molecular system of alanine dipeptide. We obtain a small set of samples from the explored part of the configuration space (expanded by the dihedral angles $\phi$ and $\psi$ in this case). With those,  we can train our dynamic surrogate model to predict the full energy landscape,  speeding up the sampling of unexplored configurations.

The high diversity of the simulation methods and the excellent results achieved in all is a strong indicator that our pipeline can be integrated across all material science applications and significantly expedite the phenomena investigation. 

% Finally, we investigate the latent space distribution created by our model to showcase the physically sound embedding achieved out of the box by our VAE. Additionally, the specific architecture chosen for modeling the dynamics allows for introspection into our machine-learning algorithm. With this, we show that the model learns physically correct parameter dependencies, which provides an a posteriori explanation for the excellent prediction results and proves that the model learns physically correct input-output relations.
\subsection{Methods}
In this section, we present the surrogate model architecture. Generally, the main goal of any ML or surrogate model is to be as widely applicable as possible. Therefore, the surrogate is trained to autoregressively model the field of interest the cell at each time step $t$,  $c_t^l(x,z)$ for a large set of simulation parameter combinations $l \in \{P\}$. Hence, the task of our surrogate model $\mathcal{S}$ can be defined as
\begin{equation}\label{eq:framework1}
    c_{t+1:t+\tau}^l(x,z) = \mathcal{S}(c_{t-k:t}^l(x,z), \Lambda^l)
\end{equation}
where $\Lambda^l$ is a vector containing a specific combination $l$ of input parameters to the dynamical simulation. $k$ is the length of the sequence of past inputs used to predict the future sequence of concentrations with length $\tau$.

\begin{figure*}[!ht]
    \centering
    \includegraphics[width=\textwidth]{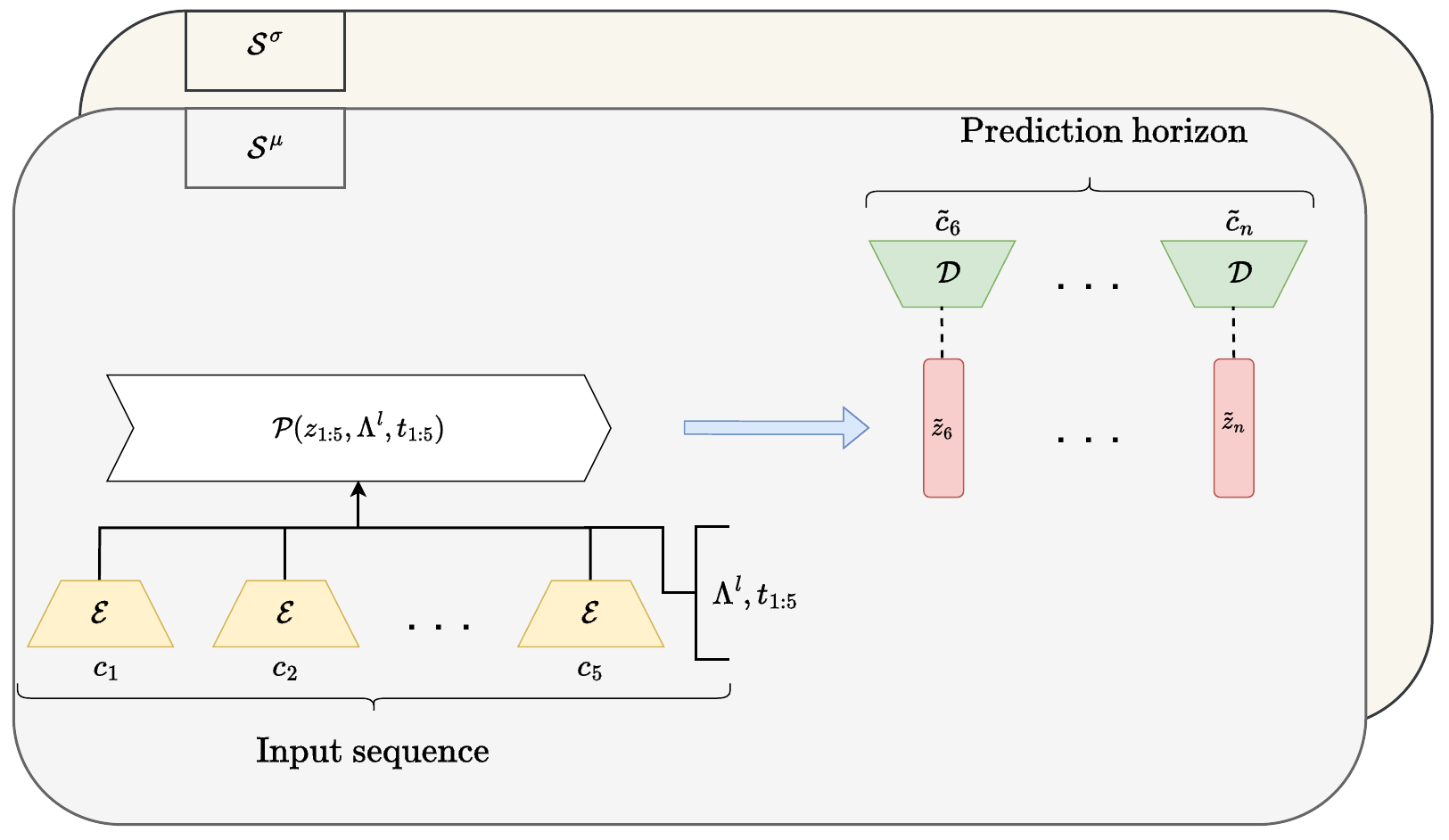}
    \caption{Visualization of the entire surrogate model using a multi-horizon forecasting model as the propagator $\mathcal{P}$. $\mathcal{E}$ and $\mathcal{D}$ are the encoder/ decoder parts of the VAE, $c_i$ are the input configurations and $\tilde{c_i}$ are the predicted next configurations. This process is repeated to predict sequences that are longer than the prediction horizon of the multi-horizon model. All the models have a $S^{\mu}$ branch that predicts the evolution of the dynamics. Stochastic models have an additional $S^{\sigma}$ branch that predicts the stochastic behaviour of the evolution.}
    \label{fig:rec_model_vis}
\end{figure*}

The idea for our model is to split the task of $\mathcal{S}$ into three subtasks and tackle each one with a specific machine-learning algorithm. \autoref{fig:rec_model_vis} illustrates the three tasks. We need an encoder $\mathcal{E}$, a propagator $\mathcal{P}$, and finally a decoder $\mathcal{D}$. $\mathcal{E}$ and $\mathcal{D}$ allow us to go from the physical space to the latent space and vice versa. The propagator $\mathcal{P}$ is specifically chosen such that it considers the sequential nature of our time series data, taking one physical state and propagating it to the next state. Traditional propagators output the next state exclusively in a recurrent manner. Our model however outputs multiple future states simultaneously with one call. This multi-horizon forecasting as this leverages the commonality expected in the subsequent steps. Succinctly the combination of the three components can be seen in \cref{fig:rec_model_vis} and \cref{eq:framework2} :

\begin{equation} \label{eq:framework2}
    c_{t+1:t+\tau}^l(x,z) = \mathcal{D}[\mathcal{P}(\mathcal{E}(c_{t-k:t}^l(x,z)), \Lambda^a)] .
\end{equation}

% In order to capture the probabilistic nature of the physical simulation, two separate but architecturally identical models are trained in parallel to predict the average property maps and the respective standard deviation across the domain at each time step. 
% For this, we train one model on the mean of the parallel trajectories for one parameter configuration. A second model is trained on the standard deviation across the same trajectories.

\FloatBarrier
\subsection{Encoder \texorpdfstring{$\mathcal{E}$}{E} and Decoder \texorpdfstring{$\mathcal{D}$}{D}}
For parametrizing our encoder $\mathcal{E}$ and decoder $\mathcal{D}$, we chose a Variational Autoencoder (VAE)\cite{kingma2013auto}. This allows for efficient dimensionality reduction while capturing the essential features of the data and thereby reducing the complexity of the task for the propagator. Additionally, it also functions as an embedding, placing physically similar configurations close to each other. 
% In \autoref{fig:my_vae}, we see an indicative architecture used for the VAE, a symmetric model with convolutional and transposed convolutional layers. It does not use any pooling layers, only Exponential Linear UNits (ELU) activation functions \cite{clevert2015fast} to enable non-linear dimensionality reduction. In the final encoding layers, the output is flattened and processed by two fully connected layers, bringing the dimension of the data down to the desired latent space dimension. 

\begin{figure*}[h!]
    \centering
    \includegraphics[width=\textwidth]{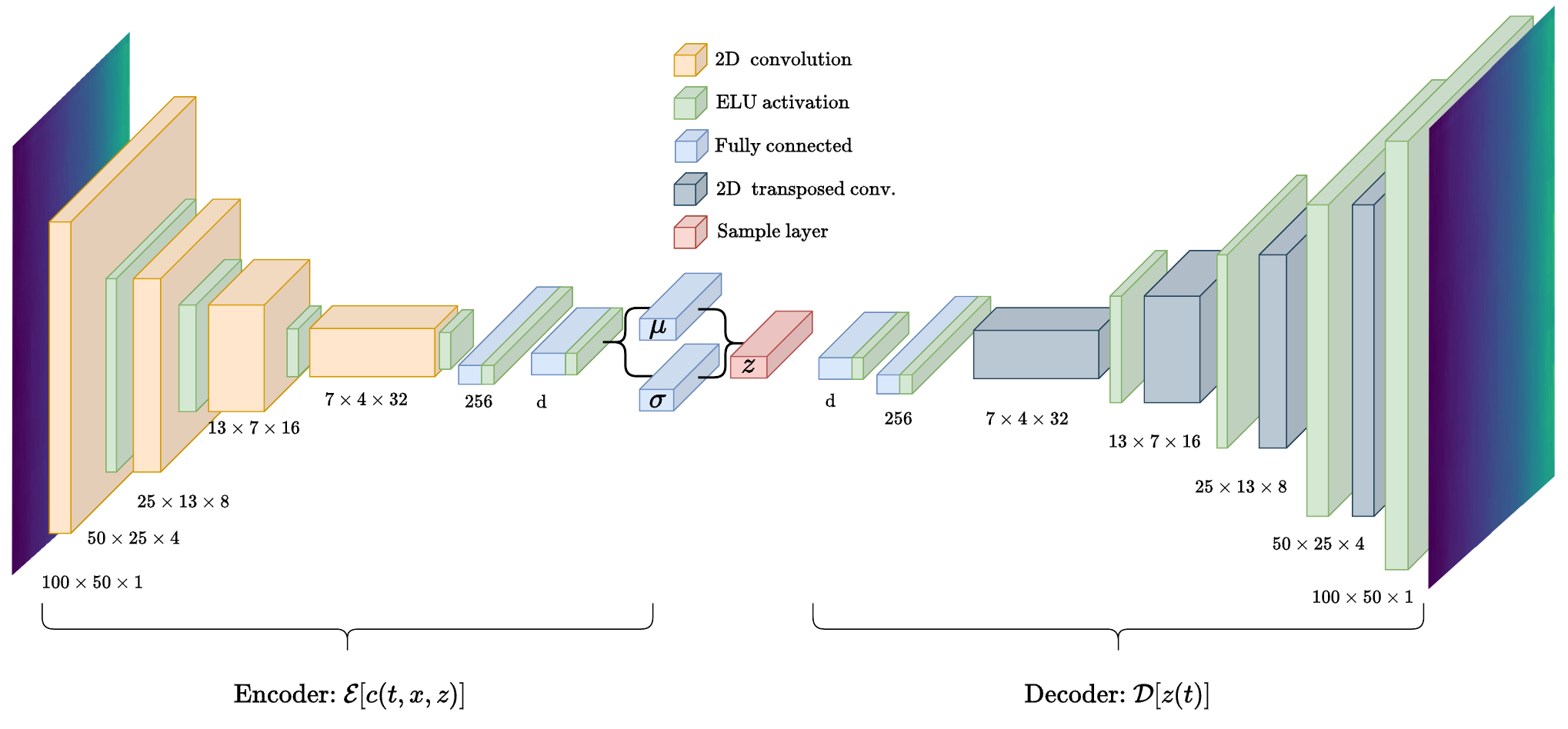}
    \caption{Schematic representation of the encoding-decoding process. A high dimensional map $c_{t_0}(x,z)$ at a given time point $t_0$ is taken and encoded onto its respective latent space representation $z_{t_0}$. It can then be decoded by propagating it through the decoder network $\mathcal{D}$, recovering the high dimensional representation $\tilde c_{t_0}(x,z)$, which should be as close to the original input as possible.}
    \label{fig:my_vae}
\end{figure*}
\FloatBarrier
The VAE is trained separately from the propagator algorithm. This design choice was made as it eases the testing and comparison of different architectures for $\mathcal{P}$, increases the modularity of the model, and finally, gives improved results. The loss, therefore, is geared towards training the model to encode and decode the same configuration as accurately as possible while producing a regularized latent space and can be concretely seen in \cref{eq:loss_vae}

\begin{equation}
\label{eq:loss_vae}
    \mathcal{L}(c, \tilde c) = \frac{1}{N_{tot}} \sum_{i=1}^{N_{tot}} (c-\tilde c)^2 - \frac{1}{2}\sum_{i=1}^{d} (1+\log(\sigma_i^2) - \mu_i^2 - \sigma_i^2). 
\end{equation}
Here $N_{tot}$ is the total number of configurations to encode. These consists of all parameter combinations $l\in \{P\}$ at all times steps $t \in \{T\}$. Two terms make up the total loss term. The first expression optimizes the reconstruction accuracy  of the output $\tilde{c}$ with respect to the input $c$ of via the Mean-squared error loss. The second forces the latent space mapping $q_x(z){\mu_i, \sigma_i}$ to be close to a standard normal $N(0,1)$ distribution. In our model, we chose to weigh both loss contributions equally as they are on a similar scale, and this provides good accuracy and regularization.
\label{sec:propagator}
\subsection{Temporal Fusion Transformer}
\label{sec:tft_theory}

% In summary these mechanisms equip the model with three key advantages compared to other models for our application case:
% \begin{enumerate}
%     \item Ability to use static and dynamic metadata i.e., simulation parameters
%     \item the learning of short- and long-term dependencies
%     \item A posteriori insight into variable importance 
% \end{enumerate}

The latent space propagator chosen is the Temporal Fusion Transformer(TFT)\cite{Lim2021} which was specifically developed for time series forecasting. It takes as input the reduced dimension latent space of the VAE and predicts its evolution over time. If the full configuration is desired, it can be re-decoded via the decoder element of the VAE.
This model is superior to models used in similar applications because of its
\begin{enumerate}
    \item Ability to use static and dynamic metadata i.e., simulation parameters
    \item the learning of short- and long-term dependencies
    \item A posteriori insight into variable importance 
\end{enumerate}

% Previous work has shown that Transformer-based models could be especially suited for this as they can be shown to be more expressive than any numerical time integration scheme \cite{Geneva2022}.  It was developed as a multi-horizon forecaster that can use mixed interrelated inputs.
% Multi-horizon means the model predicts the target always at a range of future time points. 
It can accept mixed inputs, specifically static, past, and future covariates that are taken into account in addition to the standard input to make predictions, such as time series with high correlation to the prediction target. Static covariates are quantities that give context to an input sequence but remain unchanged over time, e.g., the values of the conserved properties of our simulation. Past and future covariates serve a similar purpose but are dynamic in time. Past covariates are only known up to the prediction starting point.

% \begin{figure}[!ht]
%     \centering
%     \includegraphics[width=\columnwidth]{figures/02_methods/covariates_vis.pdf}
%     \caption{Visualization of the three different types of time series data that can be utilized with the TFT model. Multiple independent covariates series can also be used.}
%     \label{fig:cov_vis}
% \end{figure}

Several key features of the TFT architecture allow it to learn the complex relations between these mixed input signals. We will give a short introduction to these main features but the reader is encouraged to read \cite{Lim2021} for more details.

\begin{figure*}[!h]
    \centering
    \includegraphics[width = \textwidth]{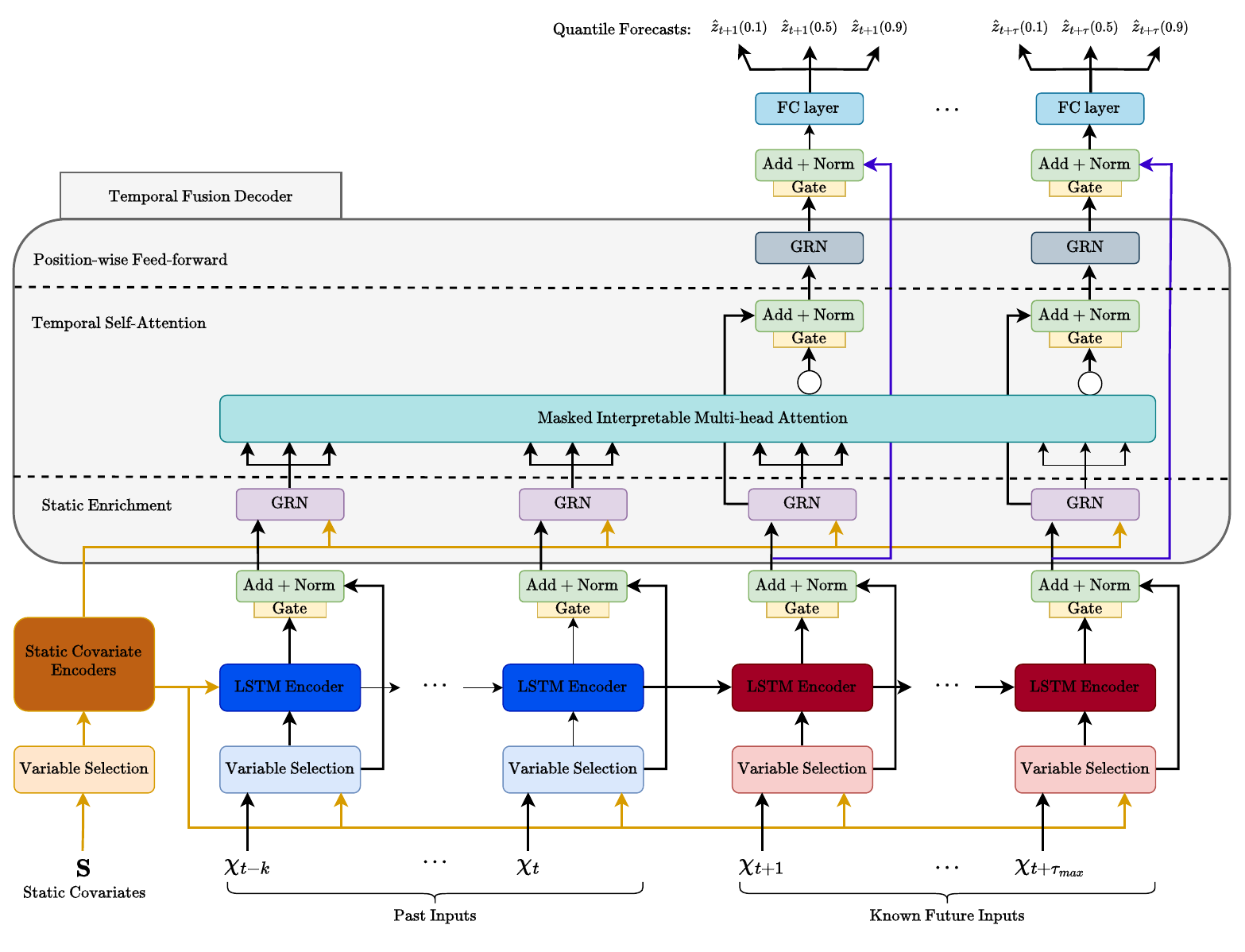}
    \caption{Temporal Fusion Transformer model, adapted from \cite{Lim2021}
    Static covariates S and time-varying inputs — past observations ($x_{t-k}...x_t$) and known future covariates ($x_{t+1}...x_{t+\tau_{max}}$)  enter through dedicated Variable Selection Networks (orange), which learn instance-wise feature weights for interpretability. The Static Covariate Encoder (dark orange) produces four context vectors used throughout the network: one for variable selection, two to initialize the LSTM's cell/hidden state, and one for static enrichment. A sequence-to-sequence LSTM Encoder–Decoder (blue = past, dark red = future) provides local temporal processing, replacing standard positional encoding. Every sub-block is wrapped in a Gate + Add and Norm unit (yellow/green), a GLU-based residual gate that lets the network skip unused components. LSTM outputs are enriched with static context via a Static Enrichment GRN (purple), then passed to Masked Interpretable Multi-Head Attention (teal), which lets each time step attend to relevant past and future points while preserving causal masking. This is followed by a Position-wise Feed-Forward GRN, another gated residual block, and a final FC layer producing quantile forecasts ($\tilde{z}_{t+\tau}$(0.1), (0.5), (0.9)) at each horizon step — enabling calibrated uncertainty estimates alongside point predictions. }
    \label{fig:tft_vis}
\end{figure*}

\textbf{Variable Selection Network:} Most ML models are black box functions that allow for little to no physical interpretation. Even though TFT is ultimately a highly complex, it  allows for relative variable importance weighting by introducing a Variable Selection Block.  The color coding in \autoref{fig:tft_vis} for the Variable Selection blocks indicates that for each type of input, i.e., static, past, and future, an individual block is used. Continuous input variables are linearly transformed into a $d_{model}$ dimensional vector before being processed by the Variable selection cell. 
% For a flattened vector of all past inputs $\Xi_t = [\xi_t^1, ..., \xi_t^{m_{\chi}}]^T$ at time $t$ containing the transformed input of each variable $j$ as $\xi_t^j \in \mathbb{R}^{d_{model}}$, the selection weights $v_{\chi_t} \in \mathbb{R}^{m_{\chi}}$ are computed as follows

% \begin{equation*}
%     \mathbf{v}_{\chi_t} = \text{Softmax}(\text{GRN}_{v_{\chi}}(\Xi_t, \mathbf{c}_s)).
% \end{equation*}
% $\mathbf{c}_s$ here again defines an external context vector obtained from the static covariate encoder shown in \autoref{fig:tft_vis}.  Non-linearity is added to the selection process by processing each $\xi_t^j$ with a GRN
% \begin{equation*}
%     \tilde\xi_t^j = \text{GRN}_{\tilde \xi(j)}(\xi_t^j).
% \end{equation*}
While each variable has its respective $\text{GRN}$, the weights are shared across all time steps $t$. Finally, the processed feature vectors are multiplied with selection weights and summed up.
% \begin{equation*}
%     \tilde \xi_t = \sum_{j=1}^{m_{\chi}} v_{\chi_t}^j \tilde\xi_t^j.
% \end{equation*}
% Analogous operations take place for future and static inputs.

\textbf{Static Covariate encoders}
The static covariate encoders denoted by the orange box in \autoref{fig:tft_vis} are specifically designed to allow the model to utilize contextual metadata throughout the prediction process. Four unique context vectors $\mathbf{c}_s, \mathbf{c}_e, \mathbf{c}_c$ and $\mathbf{c}_h$ (orange arrows in \autoref{fig:tft_vis}) are passed into the model for (1) Temporal variable selection ($\mathbf{c}_s$), (2) initialization of the LSTM encoders for local processing ($\mathbf{c}_c,\mathbf{c}_h $) and (3) to enrich the temporal features with static information in the decoder block ($\mathbf{c}_e$). GRN cells are used to create these context vectors.\\

\textbf{Long Short Term Memory (LSTM) Encoders}
Short range dependencies are captured by LSTM blocks \cite{hochreiter1997long}, which share inputs derived from the static covariate encoder and the variable selection. The blocks processing past covariates share weights with each other. Equally, the blocks processing future known covariates share weights as well, as indicated by the colour coding of \cref{fig:tft_vis}.

\textbf{Gating Mechanism}
A Gated Residual Network (GRN) is introduced to allow the weighing of different input variables and switching between linear and non-linear processing. This block processes the primary input and an optional context vector:

% \begin{align*}
%  \text{GRN}_{\omega}(\mathbf{a,c}) &= \text{LayerNorm}(\mathbf{a}+\text{GLU}_{\omega}(\mathbf{\eta}_1))\\
%  \mathbf{\eta}_1 &= \mathbf{W}_{1,\omega}\eta_2 + \mathbf{b}_{1,\omega}\\ 
%  \mathbf{\eta}_2 &= \text{ELU}(\mathbf{W}_{2,\omega} \mathbf{a}+ \mathbf{W}_{3,\omega} \mathbf{c} +\mathbf{b}_{3,\omega}).
% \end{align*}
LayerNorm is the standard layer normalization introduced in \cite{ba2016layer}.Weights are shared across the layer, in \autoref{fig:tft_vis} as is indicated by the color coding of the different GRN blocks. Component gating layers are used, which are based on Gated Linear Units (GLUs) \cite{dauphin2017language}. 
% An input $\gamma$ is then processed as follows
% \begin{equation*}
%     \text{GLU}_{\omega} = \sigma(\mathbf{W}_{4,\omega} \gamma +\mathbf{b}_{4,\omega}) \odot (\mathbf{W}_{5,\omega} \gamma +\mathbf{b}_{5,\omega}).
% \end{equation*}
% Here  $\sigma(\cdot)$  is the sigmoid activation function. 
GLUs allow the model to learn how much the GRN should modify the original input. If no context vector is passed the block accepts 0s as inputs and is organically deactivated.
% \begin{figure}[!ht]
%     \centering
%     \includegraphics[width=\columnwidth]{figures/02_methods/grn_vrn_combined.drawio.pdf}
%     \caption{The Gated Residual Network block uses the ELU activation function \cite{clevert2015fast} and fully connected (FC) layers. The Variable Selection network is made up of several independent GRN blocks. Adapted from \cite{Lim.2021}}
%     \label{fig:grn_vis}
% \end{figure}

\textbf{Interpretable Multihead Attention:} In \cite{Lim2021} a modified version of the Multi-Head attention is presented to improve model introspection. Attention weights in each head depend on the specific value weights in that head $\mathbf{W}_V^h$. Hence, they cannot be aggregated and evaluated. Therefore, a modified attention algorithm that shares values across heads is introduced

% \begin{align*}
% & \text{InterpretableMultiHead}(\mathbf{Q,K,V}) = \mathbf{\tilde HW}_H \\
%     \mathbf{\tilde H} &= \tilde A(\mathbf{Q,K})\mathbf{VW}_v\\
%     &= \biggl\{ \frac{1}{m_H} \sum_{h=1}^{m_H} A(\mathbf{QW}_K^h,\mathbf{KW}_K^h)\biggr\}\mathbf{VW}_V\\
%     &=\frac{1}{m_H} \sum_{h=1}^{m_H} \text{Attention}(\mathbf{QW}_Q^h, \mathbf{KW}_K^h, \mathbf{VW}_V)
% \end{align*}

Now the value weights are shared across all  attention heads and combined by a linear mapping. We can interpret this new form of attention as an ensemble of multiple attention heads that can attend to different temporal patterns.\\

\FloatBarrier
\section{Stochastic Methods}
% \section{Methods}
\subsection{Metropolis Monte Carlo Model for ion separation}\label{sec:ising}
The Metropolis Monte Carlo model, as described in \cite{kotze2008introduction}, treats a system as a two-dimensional grid where dipole spins are placed randomly at lattice sites. The  The spins are constrained to take one of two values: up $(+1)$ or down $(-1)$. The lattice is defined by dimensions $W \times L$, and in this example $W = L$, such that the total number of spins is $L \times L$. As a result, for a specific sites $i$, a Hamiltonian is produced 
\begin{equation}\label{eq:hamiltonian}
    \epsilon_i = -J \sum_{ j } (s_i s_j - 1)\tag{2}
\end{equation}
where the sum runs over the nearest neighbours of i-the Spin. The coupling constant J characterizes the natural interaction within the Ising Model, with its sign determining the nature of the interaction. A positive J corresponds to a ferromagnetic system, where the spins tend to align parallel to minimize energy. Conversely, a negative J signifies an antiferromagnetic system, where the spins prefer an anti-parallel alignment.
% we will illustrate the first iteration of the algorithm using a simple $3 \times 3$ spin matrix. 
We will consider a grid of spins where each spin can either be up (+1) or down (-1). The goal is to minimize the magnetic energy of the system by performing spin exchanges based on the energy difference between configurations. The deep learning pipeline is operating similarly, with TFT predicting the energy evolution of a system given its parameters and a recurrent VAE predicting the next configuration given the energy predicted by TFT. 

We use this simulation set up to investigate the spin exchange of a 2 dimensional system. We differentiate different systems by altering  the following properties of the system.

\begin{figure}[h!]
    \centering
    % First row of images
    \begin{minipage}[t]{0.24\textwidth}
      \includegraphics[width=\textwidth]{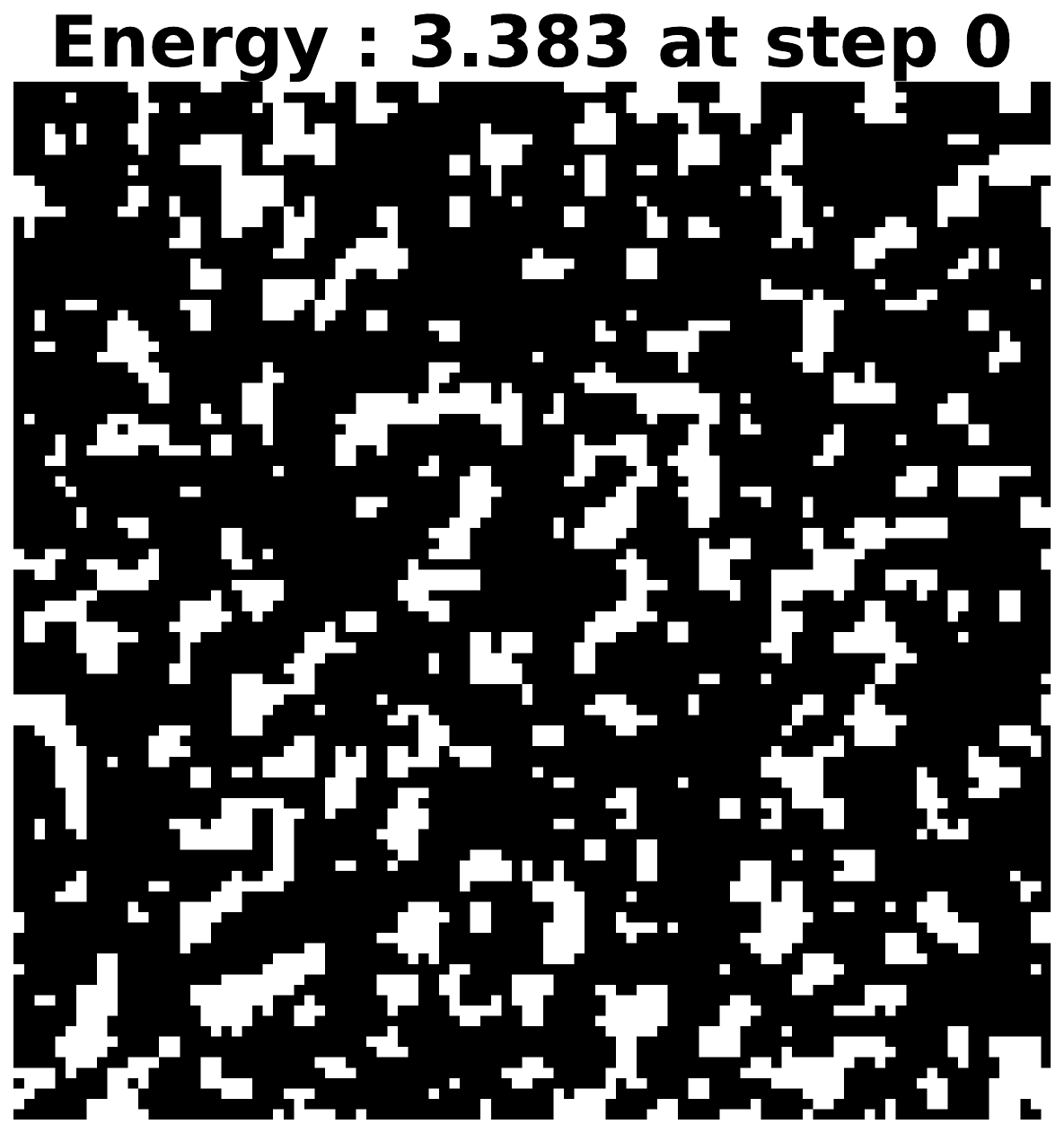}
    \end{minipage}
    \hfill
    \begin{minipage}[t]{0.24\textwidth}
      \includegraphics[width=\textwidth]{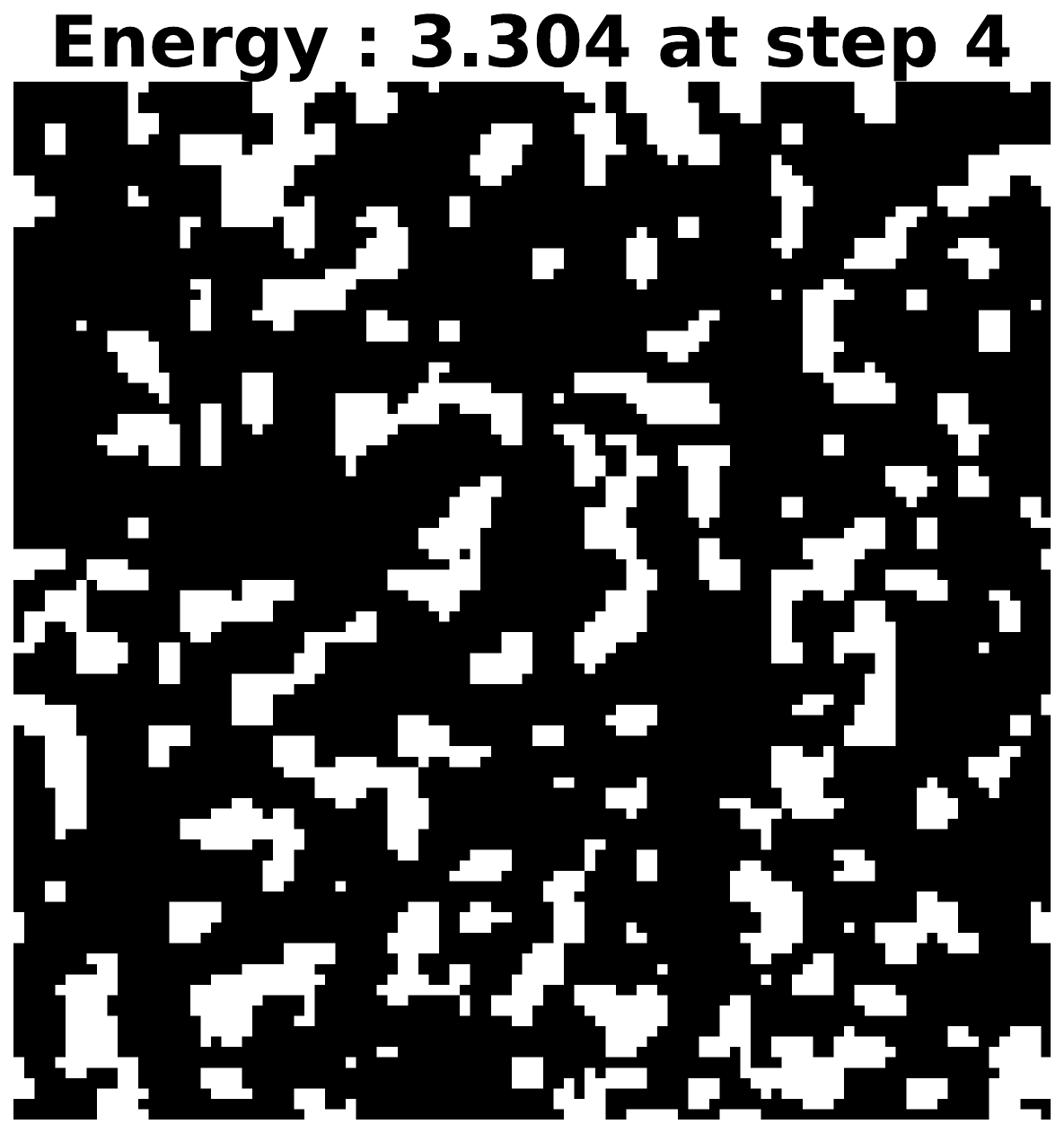}
    \end{minipage}
    \hfill
    \begin{minipage}[t]{0.24\textwidth}
      \includegraphics[width=\textwidth]{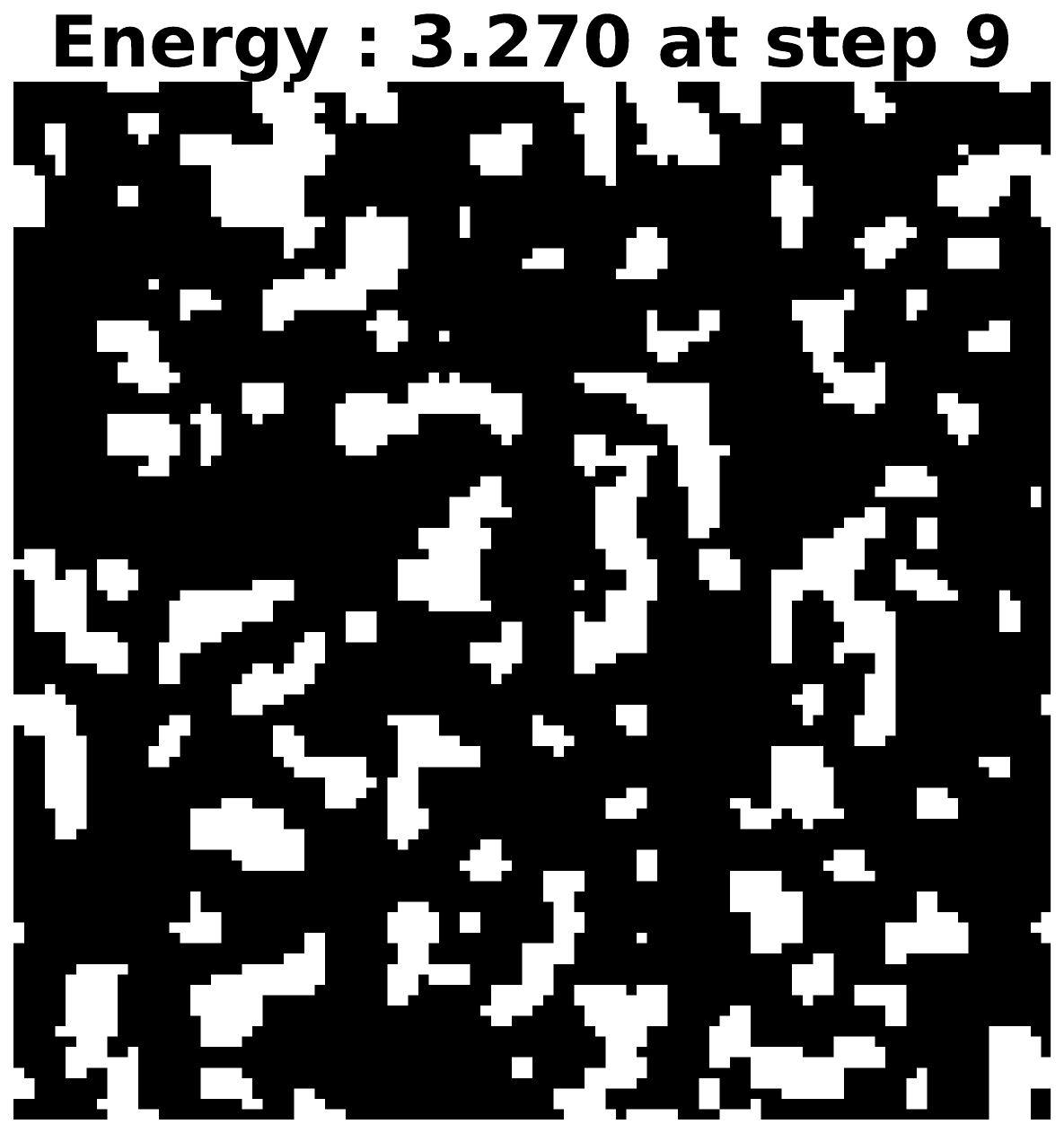}
    \end{minipage}
    \hfill
    \begin{minipage}[t]{0.24\textwidth}
      \includegraphics[width=\textwidth]{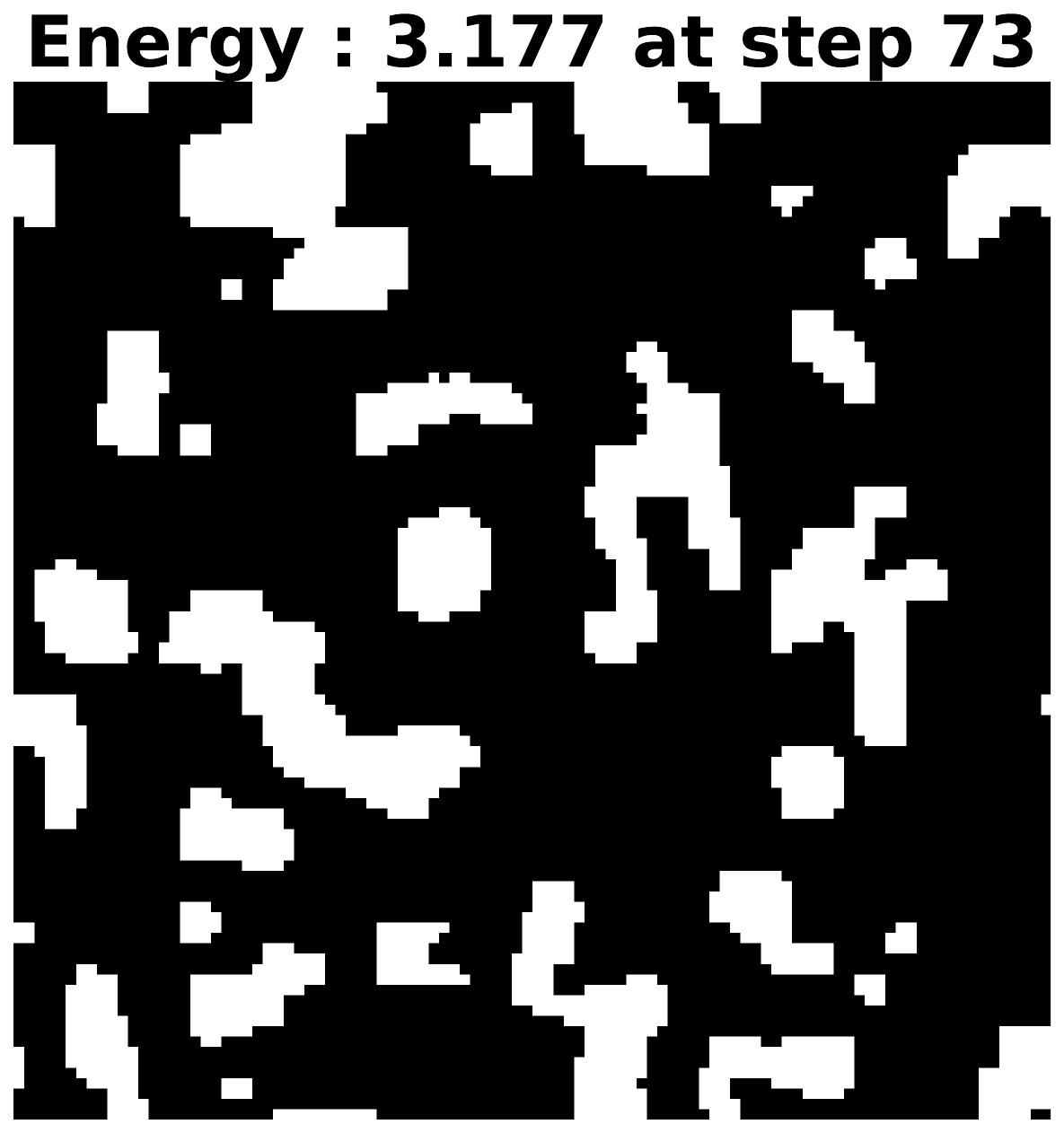}
    \end{minipage}
    
    \vspace{0.5cm} % Space between rows

    % Second row of images
    \begin{minipage}[t]{0.24\textwidth}
      \includegraphics[width=\textwidth]{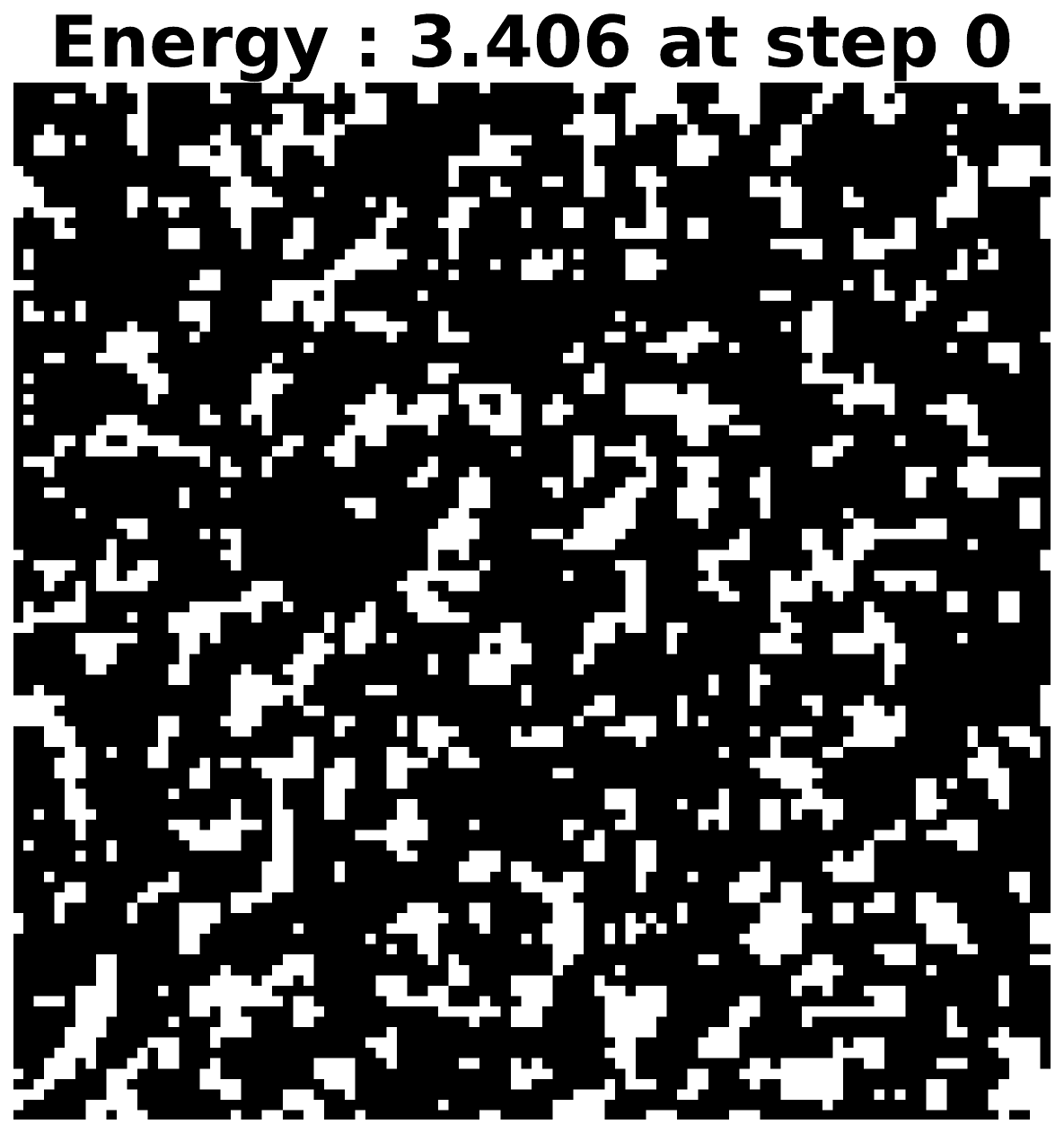}
    \end{minipage}
    \hfill
    \begin{minipage}[t]{0.24\textwidth}
      \includegraphics[width=\textwidth]{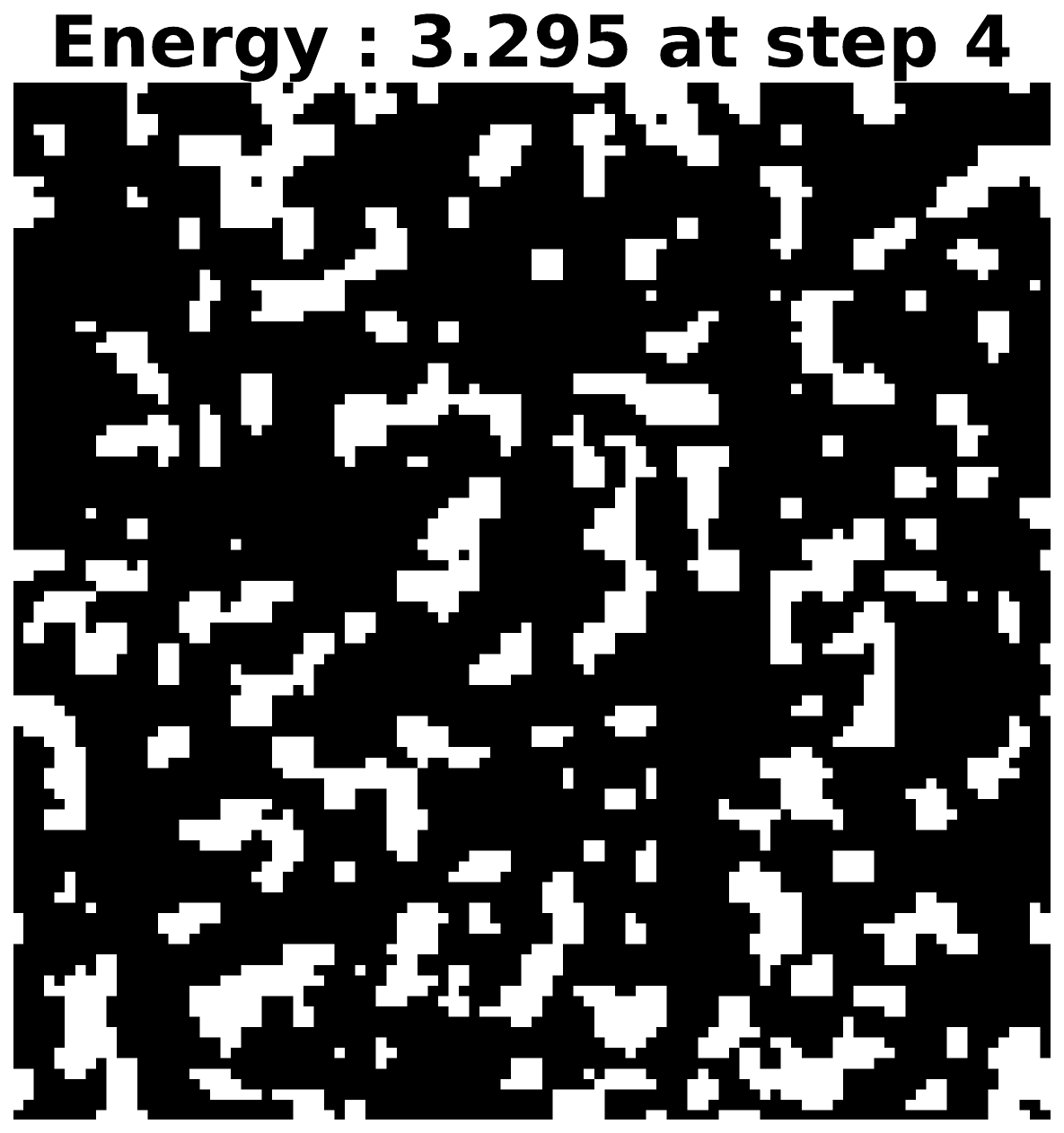}
    \end{minipage}
    \hfill
    \begin{minipage}[t]{0.24\textwidth}
      \includegraphics[width=\textwidth]{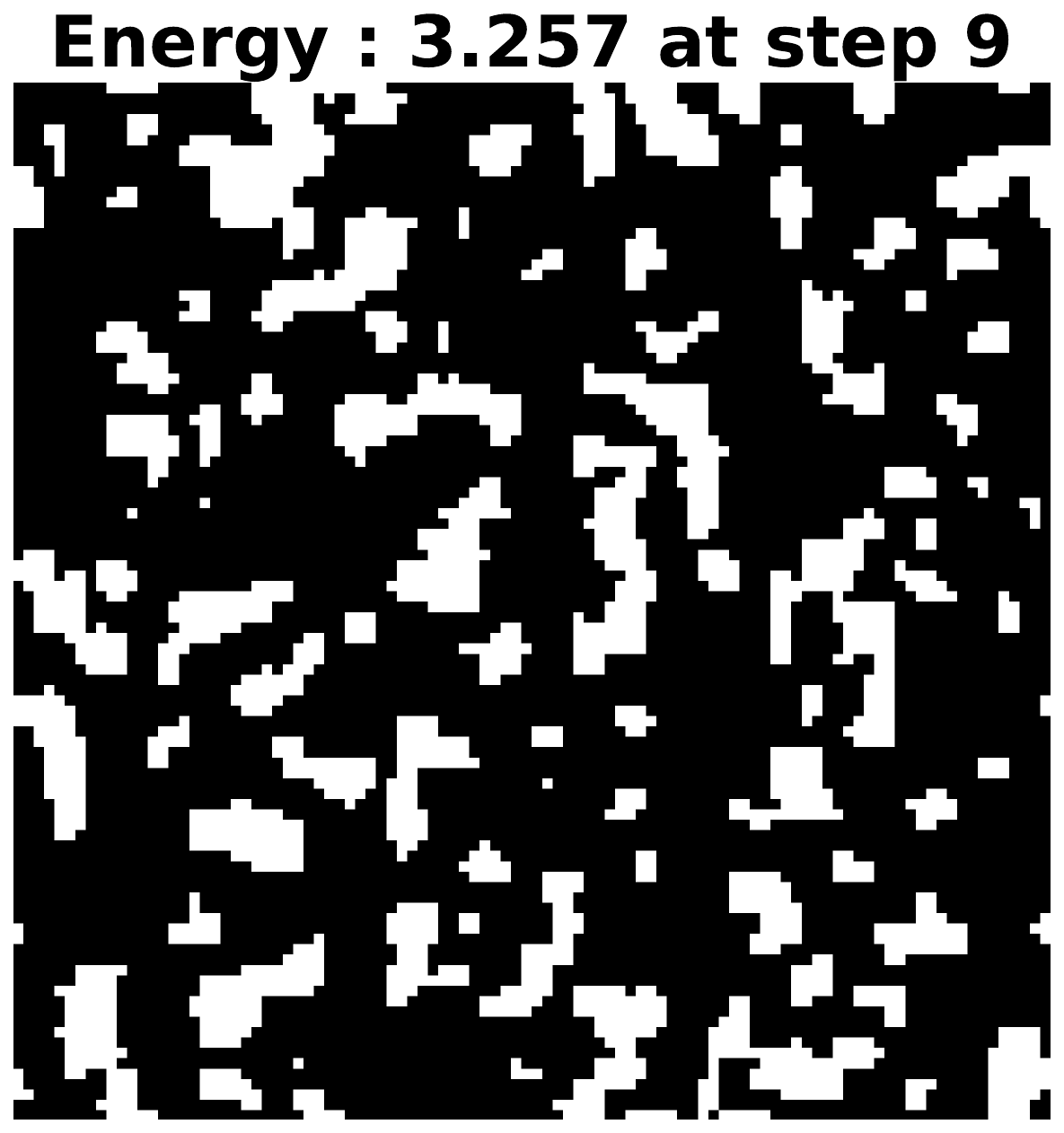}
    \end{minipage}
    \hfill
    \begin{minipage}[t]{0.24\textwidth}
      \includegraphics[width=\textwidth]{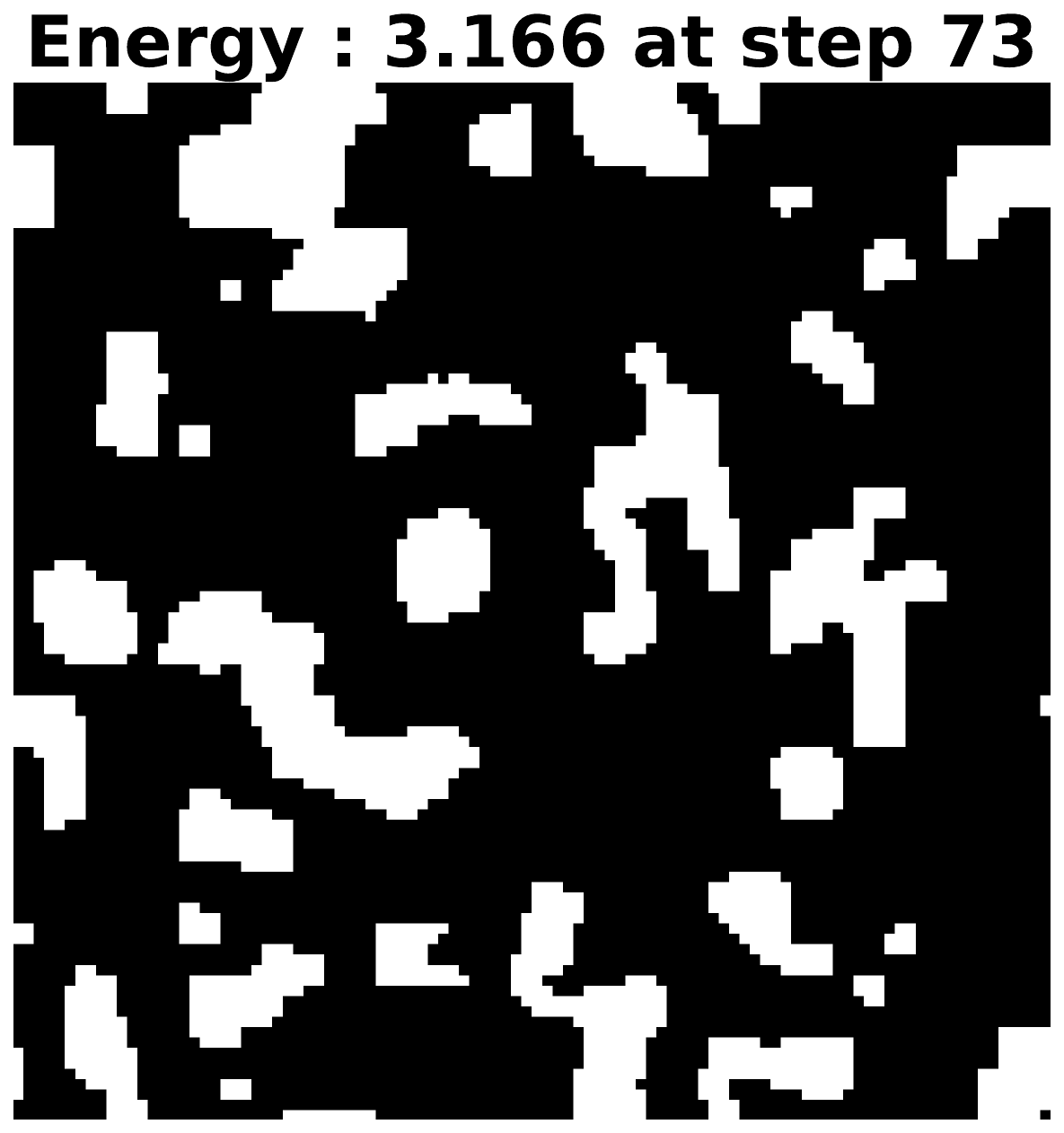}
    \end{minipage}
    \caption{Snapshots generated through the physical simulation (up) and snapshots generated through our framework (down) with their energies and corresponding timesteps}
    \label{fig:ising_results}

  \end{figure}

\begin{table}[h!]
    \centering
    \caption{Simulation Parameters, as this is an abstract case the units are irrelevant.}
    \begin{tabular}{lll}
        \toprule
        \textbf{Parameter} & \textbf{Description} \\
        \midrule
        J           & Exchange interaction energy between spins & 0.1, 0.5, 1, 2, 3, 5, 6, 7, 8 \\
        $K_{BT}$       & Thermal energy (Boltzmann constant multiplied by temperature) & 0.1, 0.5, 0.8, 1, 2, 2.5, 3, 5. \\
        Ratio       & The probability \( P(X=1) \) & 0.1, 0.25, 0.4, 0.5, 0.75, 0.8 \\
        \bottomrule
    \end{tabular}
    \label{tab:simulation_parameters}
\end{table}
To fully describe the phenomenon we need to reconstruct the energy evolution, as well as the convergent energy of each parameter combination. The results are presented in \cref{fig:ising_results} and show a high energy correlation with the ground truth ( lower than $5\%$)
% Additionally, the prediction of the standard deviation allows not only the average phenomenon reconstruction but also the introduction of a good estimate of the epistemic uncertainty of the physical model.
% \begin{figure}[h!]
%     \centering
%     \includegraphics[width=0.5\linewidth]{figures/01_theory/1_kmc_model.pdf}
%     \caption{Caption}
%     \label{fig:ising_results}
% \end{figure}
In this instance, the exact configuration of the ions is of little interest as multiple configurations will be equivalent. Nevertheless, a full visualization of the phenomenon is of significant aid to the understanding of it. 
To this end, we modified the VAE architecture described in \cref{fig:my_vae} in two significant ways. The first added physical insight to the latent space by appending the predicted energy (via TFT) and the spin ratio to it. The loss function, additionally to the reconstruction, included the MSE between the predicted and true energy and ratio of configurations. To further improve the results we modified the architecture to resemble that of a Residual Neural Network (ResNet) \cite{he2016deep}. In that architecture the predicted target is not the next step but the difference of the current step to the next. The reconstructed next step is then created by adding the output of the network to the input, resembling the analytical solution of the explicit Euler propagation scheme. As the spin values could only be between 0 and 1, we further processed the end result with a sigmoid. 
The results of the reconstructed images, along with their energies can be seen in \cref{fig:ising_results}. 

As is obvious, the method works robustly and well, and the addition of the ResNet provides a consistent foundation for the image reconstruction. 
\FloatBarrier

\subsection{Kinetic Monte Carlo for Solid State Electrolytes}
Even though the Ising model described in \cref{sec:ising} provides a good proof of concept, applications of interest require more complex simulation methods. To further test our prediction pipeline, we deployed it to a full kinetic Monte Carlo model applied in the investigation of the boundary layer formation of solid state electrolytes. We used the full configutaion as input to the VAE. The resulting latent space time series provided the time series predicted by TFT, while the simulation paramters provided the static covariates. The decoder of the VAE expanded the predicted latent space to the original configuration. 

\subsection{Kinetic Monte Carlo}
\label{sec:kMC}
The kinetic Monte Carlo method provides a numerical algorithm to model physical phenomena by coarse-graining the dynamics into a set of long-term states. This, in turn, enables, for example, the simulation of battery systems at the device scale. In the case of SSE modeling, this localized state $i$ is represented by the charge carrier distribution at a certain time. The physical knowledge about the process to be modeled is encoded into the transition rates between these localized states. They are implicitly defined a priori either from experiment or from underlying model equations. Implicitly, because often they are dependent on the local environment and therefore need to be computed dynamically during the simulation. Hence, by choosing the rates, we determine which processes are to be included in the model and which processes are coarse-grained into the long-term states. All transitions on smaller time scales are neglected as long as they do not change the long-term state \cite{Sickafus.2007}. In the current work we chose to simulate the boundary layer formation in Solid State Electrolytes (SSEs)
In general, the mass transport of Li-ions in SSEs can be captured by a thermally activated hopping mechanism between unoccupied vacancies in a crystal lattice. The crystal structure itself consists of immobile anions, cations and vacancies as well as mobile cations. The kMC simulation only considers the transport of mobile $Li^+$ within a three-dimensional regular grid of vacancies with lattice constant $a_\text{L}$. Note that the implemented grid does not resemble the actual morphology of the SSE sample but rather must be regarded as a simple lattice gas model. In this framework, SCL formation is caused by the mere redistribution of mobile $Li$-ions driven by an applied bias potential, $\phi_\text{bias}$.
The exact model and approximations taken can be found in \cite{kouroudis2023utilizing}.

\begin{figure*}[h!]
    \centering
    \includegraphics[width=\textwidth]{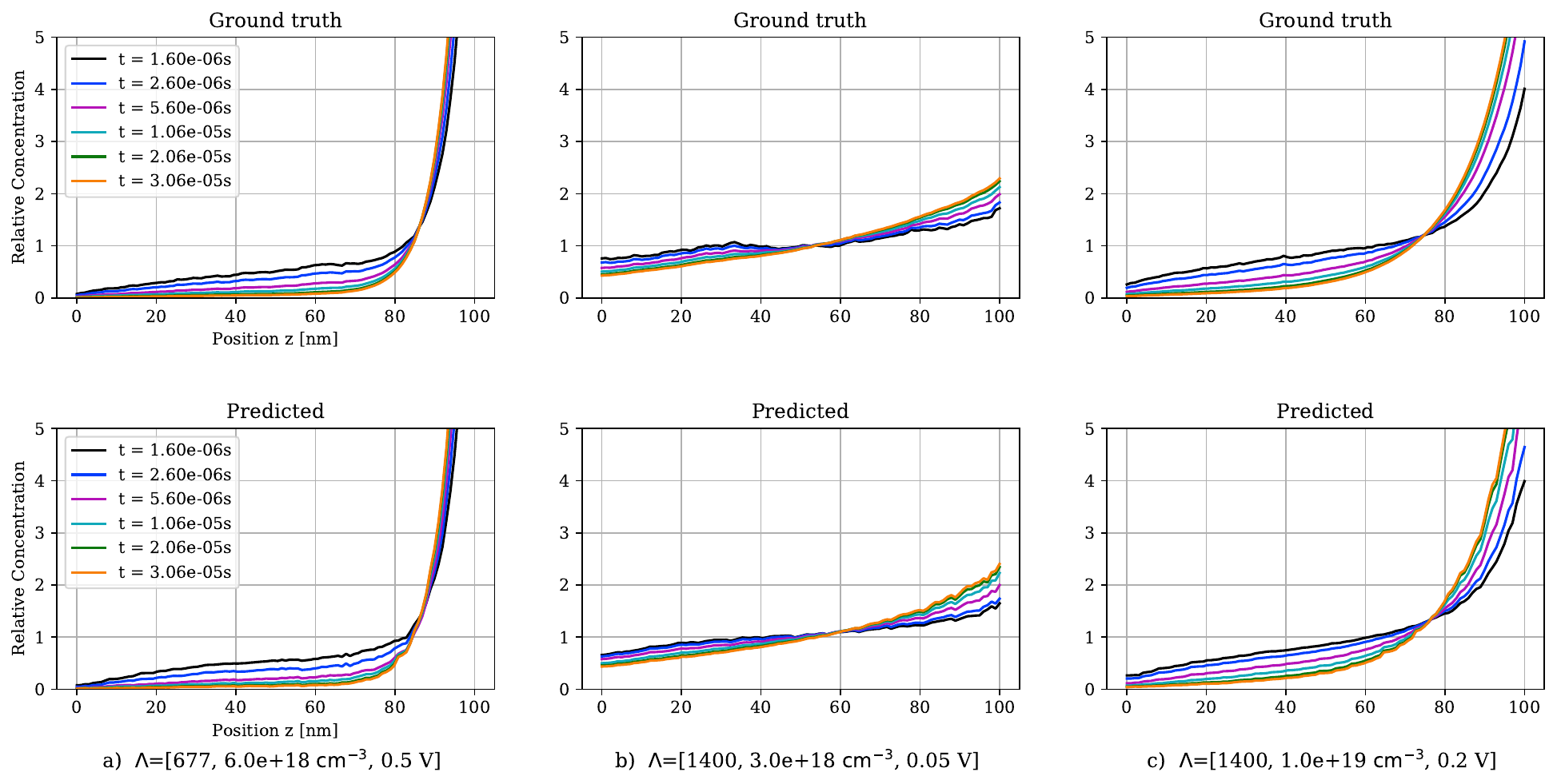}
    \caption{Predictions of the surrogate model for the mean field of the three different parameter configurations \textbf{a)}-\textbf{c}. The combination of parameters strongly affects the final concentration profile as well as the dynamic process that precedes the steady-state concentration distribution. The time stamps correspond to the following list of time steps: [10, 20, 50, 100,  200,  300]. The represetation shows the evolution of the concentration summed up along the Z axis and averaged over multiple simulation runs.}
    \label{fig:tft_1d_pred_avg}
\end{figure*}

The results can be seen in \cref{fig:tft_1d_pred_avg} and validate the robustness of our model. 

\begin{figure*}[!ht]
    \centering
    \includegraphics[width=\textwidth]{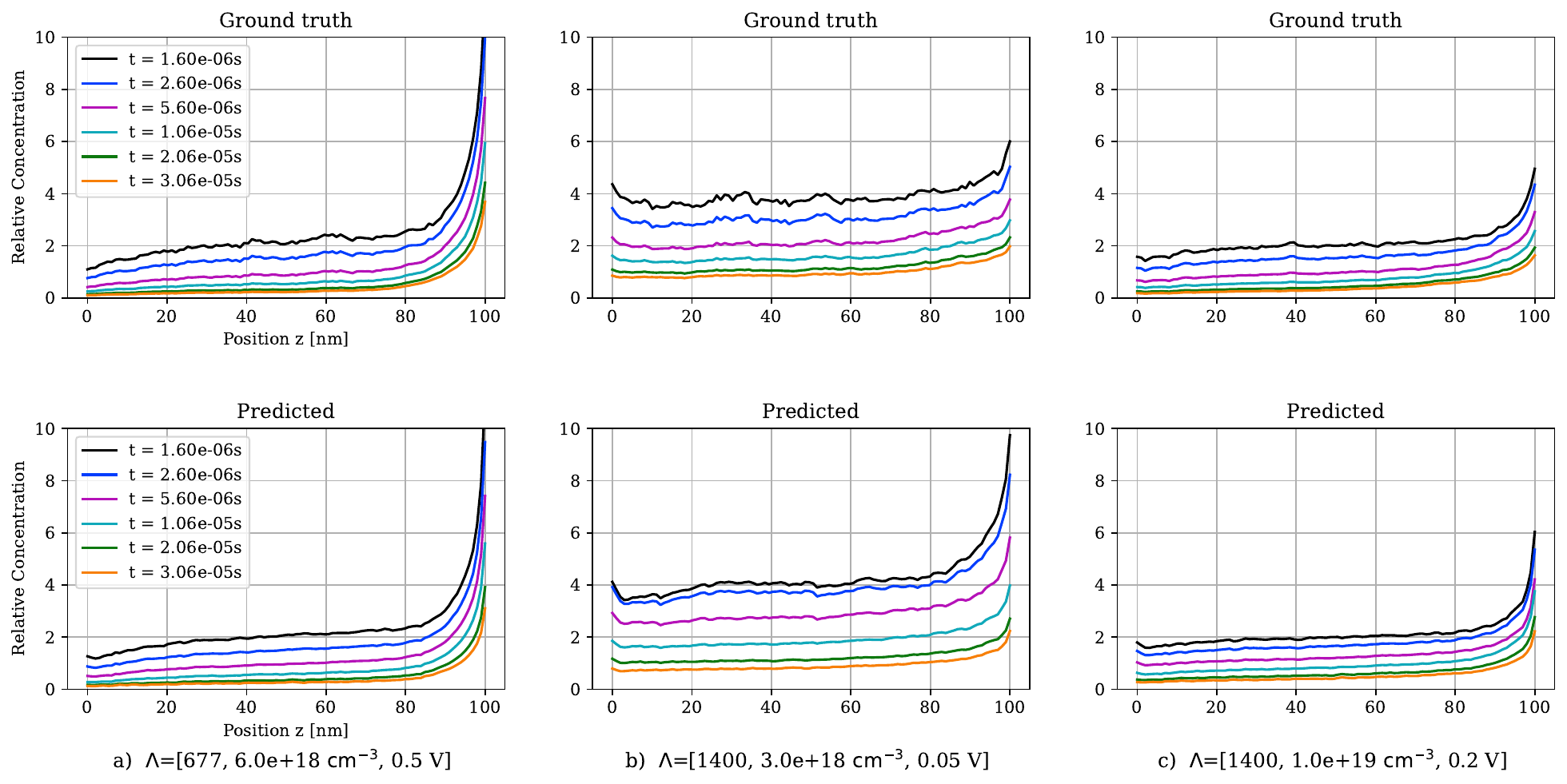}
    \caption{Prediction for the standard deviation fields of the kMC simulation. The represetation shows the evolution of the concentration summed up along the Z axis and averaged over multiple simulation runs.  Parameter configurations are the same as in \autoref{fig:tft_1d_pred_avg}.}
    \label{fig:tft_1d_pred_std}
\end{figure*}

Additionally, KMC's main advantage is its stochastic nature. To this end, we also show that our model can reclaim the underline epistemic uncertainty by predicting the standard deviation field, as shown in  \cref{fig:tft_1d_pred_std}.

% The results of the test set mean field and standard deviation field can be seen respectively seen in \cref{fig:tft_1d_pred_avg} and \cref{fig:tft_1d_pred_std}.

Furthermore, the training time of the model is in the order of minutes, while the required inference time is measured in seconds. The simulation time however spans days. This adds an additional argument in favour of our framework, namely its high accuracy is compounded by many orders of magnitude increase in efficiency.

% Lastly, leveraging the decoder part we can fully reconstruct the ion distribution as seen in \cref{fig:kmc_results}
% \begin{figure}[!ht]
%     \centering
%     \includegraphics[width=1\linewidth]{figures/03_results/kmc_results/vae_2d_paperavg.pdf}
%     \caption{Indicative results for reconstructed ion distribution (above) and the true average distribution (bottom) at time step 50, 200 and 499 respectively.}
%     \label{fig:kmc_results}
% \end{figure}
\FloatBarrier
\section{Non Stochastic Methods}
\subsection{Dynamics of Molecular Systems}
To design molecular systems in an effective manner, their configurational space needs to be explored in a robust way. Atomistic simulations using density functional theory (DFT) are regularly used in design exploration but this level of theory is limited in time and spatial scalability. Molecular dynamics use computationally efficient empirical equations of motion, parameterized usually using DFT data, to break this scalability barrier by yielding accuracy and the ability to calculate electronic properties. This offers huge possibilities to study dynamics of large complex systems. Nevertheless as the systems under study get larger and more complex, this method also becomes computationally prohibitive. One area of research for the acceleration of dynamics is the consideration of metastable states in which the system can become stuck in a local minima until a rare event leads to an escape. The dynamics inside the metastable basin, although of little consequence to the conformational dynamics, need to be sampled for a large number of steps. As the next step in the development of our framework, we applied it as a surrogate model for dynamics of a molecular system. To this end, we chose a coarse grained model of alanine dipeptide. It has a well defined and widely studied conformational space spanned by the two dihedral angles $\phi$ and $\psi$.
\begin{figure*}[!ht]
    \centering
    \includegraphics[width=0.8\textwidth]{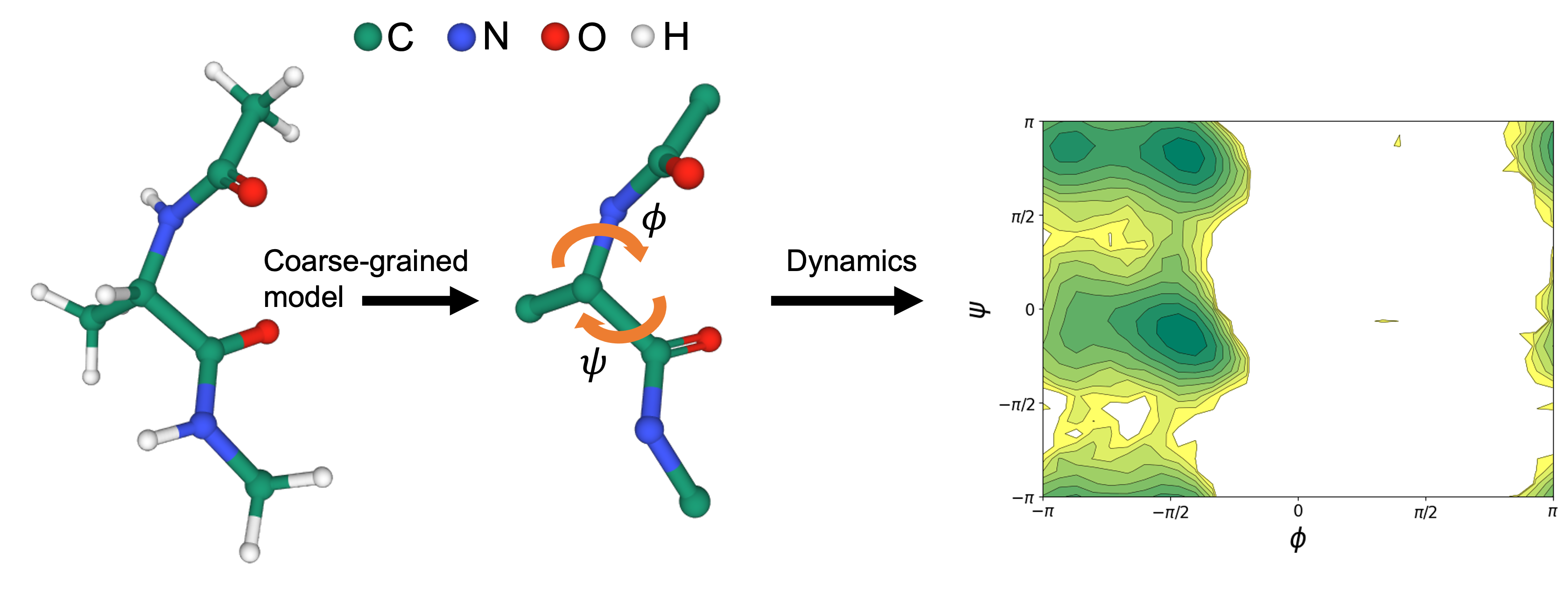}
    \caption{The system used for surrogate of molecular dynamics. The hydrogens are removed from the MD configurations. The two ramachandran angle $\phi$ and $\psi$ are indicated.}
    \label{fig:ala2_system}
\end{figure*}
\subsection{Latent dynamics}
%To reduce the dimensionality of the dynamics, we first isolate the alanine dipeptide from the explicit water and then remove all the hydrogen atoms. This is done because 1. the hydrogens undergo very high frequency vibrations not relevant to the conformational dynamics and 2. they can be added back after to the configurations output of the surrogate model by hydrogen addition rules [REF]. \\
We ignore the hydrogen atoms due to their high frequency vibrations and negligible effects on the conformational dynamics. To produce a latent embedding dynamics, we use a bond-graph architecture inspired by \cite{dobers2023latent} to focus on the topological information. 
The encoder $\mathcal{E}$ takes as input a graph $\mathcal{G}^E \in \{\mathcal{V}^E, \mathcal{B}^E\}$ that is constructed by the internal coordinates of the molecular conformations. We use bonds as vertices featurized with atomic numbers of the two atoms and the distance between them $\mathcal{V}^E = \mathbb{B}_i \sim (a_0, a_1, d_i)$ where $i$ denotes the dependence on considered frame (time step). Edges represent either angles or torsion angles featurized by the value of the respective angle $\mathcal{B}^E = \mathbb{A}_i \sim (1, 0, a_i) + \mathbb{T}_i \sim (0, 1, t_i)$. $\mathbb{B}_i$ denotes bonds in frame $i$, $\mathbb{A}_i$ the angle created by pairs and $\mathbb{T}_i$ by triplets of bonds. The encoder $\mathcal{E}$ acts on the graph $\mathcal{G}^E_i$ by first embedding the scalar features by a set of learnable MLPs (one for each feature) consisting of one hidden layer of length $H_e$ to compute the initial embeddings $\mathbf{h}^0_a$ for each node $a$ and $\mathbf{h}^0_b$ for each edge $b$. The steps of the encoding are then as follows;
\begin{enumerate}
    \item $L$ message passing steps similar to \cite{shi2021masked}:
    \begin{align*}
        \alpha_{ab} &= softmax (\frac{(\mathbf{W}_3 \mathbf{h}_a^l)^T)(\mathbf{W}_4 \mathbf{h}_b^l + \mathbf{W}_6 c_{ab}}{\sqrt{H_e}})
        \\
        \mathbf{m}_a &= \sum_{b \in \mathcal{N(a)}} \alpha_{ab} (\mathbf{W}_2 \mathbf{h}_b^l + \mathbf{W}_6 c_{ab}) \\
        \beta_a &= sigmoid(\mathbf{W}_5[W_1 \mathbf{h}_a^l, \mathbf{m}_a, \mathbf{W}_1 \mathbf{h}_a^l - \mathbf{m}_a])
        \\
        \mathbf{h}_a^{l+1} &= \beta_a \mathbf{W}_1 \mathbf{h}_a^l + (1 - \beta_a) \mathbf{m}_a
    \end{align*}
    where $W_*$ are learnable parameters, $H_e$ is hidden size of attention heads, $[a,b]$ represent vector concatenation of $a$ and $b$, $c_{ab}$ are the edge features of edges $a$ and $b$ and $\mathcal{N}(a) = \{b \mid (a,b) \in \mathcal{B} \}$. $ELU$ nonlinearities and batch normalization is applied between each layer.
    \item Pooling is done via a learnable set-to-set mapping using an $LSTM$ layer:
    \begin{align*}
        \mathbf{q}_0 &= [0...0]^T
        \\
        e_{a,t} &= h_a^L \cdot \mathbf{q}_t
        \\
        \gamma_{a,t} &= \frac{\exp(e_{a,t})}{\sum_b exp(e_{b,t})}
        \\
        \mathbf{r}_t &= \sum_{a = 1}^{N} \gamma_{a,t} h_i^L
        \\
        \mathbf{q}_t^* &= [\mathbf{q}_t, \mathbf{r}_t]
        \\
        \mathbf{q}_{t+1} &= LSTM(q_t^*)
    \end{align*}
    Where $\cdot$ denotes dot product. We perform $T$ aggregations steps
    \item Final linear layer, the latent embedding for frame $i$, $\mathbf{z}_i$ with length $L_d$ is obtained by:
    \begin{equation*}
        \mathbf{z}_i = \Phi(q_T^*)
    \end{equation*}
\end{enumerate}
The encoding process of bonds $\mathbb{B}_i$, angles $\mathbb{A}_i$ and torsion angles $\mathbb{T}_i$ for frame $i$ can be represented by the following equation
\begin{equation*}
    \mathbf{z}_i = \mathcal{E}(\mathcal{G}_i^E), \quad \mathcal{G}_i^E = \mathcal{G}^E(\mathbb{B}_i, \mathbb{A}_i, \mathbb{T}_i)
\end{equation*}

For the decoder model, we use another graph neural network with nodes encoding atomic species and edges encoding neighbor atoms. We only include time-invariant topological information in this graph which is processed via $L$ message passing steps and combined with the latent space vector $\mathbf{z}_i$ to obtain the time-dependent topological variable (bond lengths, angles and dihedrals).
The decoder graph $\mathcal{G}^D \in \{\mathcal{V}^D, \mathcal{B}^D\}$ is obtained with the following components;
\begin{itemize}
    \item $\mathcal{V}^D$ is a concatenation of time-invariant variables;
    \begin{enumerate}
        \item Scalar features: atomic number, bond degree, number of rings the atom is involved in, implicit valence, formal charge, number of bonded hydrogens
        \item Categorical features: chirality, hybridization type, is it in an aromatic ring, is it in a 5-ring, is it in a 6-ring, the name of the residue
    \end{enumerate}
    Categorical embedding is used for categorical features. The categories are based on the $enum$ structures defined in \textit{RDkit} library.
    \item $\mathcal{B}^D$ contains all the bond-neighbors in the molecule. In addition we extend the connections between any atoms that can be reached with a maximum of $k$ hops. Thus $\mathcal{B}^D = \{(a,b) \mid a \in \mathcal{V} \wedge b \in \mathcal{N}^k(a) \}$, where $\mathcal{N}^k(a)$ represent the up to $k$-hop neighbors. The edges are featurized with a categorical variable defining the type of connection (\textit{single}, \textit{double}, \textit{triple}, \textit{aromatic}, \textit{virtual}) with \textit{virtual} as the category of more than 2-hop neighbors. The edges between bonded atoms additionally have the equilibrium bond distance in the feature vector which is zero in the other edges.
\end{itemize}
The decoder $\mathcal{D}$ contains the same attention-based message passing steps as $\mathcal{E}$. After $L$ steps, we use the node embeddings $\mathbf{h}_a^L$ along with the latent space vector $\mathbf{z}_i$ to get the predictions of the time-dependent topological variable;
\begin{align*}
    d_{ab}^i &= \Gamma_{bond} ([\mathbf{h}_a^L, \mathbf{h}_b^L, \mathbf{z}_i]) \forall (a,b) \in \mathbb{B}
    \\
    \phi_{abc}^i &= \Gamma_{angle} ([\mathbf{h}_a^L, \mathbf{h}_b^L, \mathbf{h}_c^L, \mathbf{z}_i]) \forall (a,b,c) \in \mathbb{A}
    \\
    \cos \psi_{abdc}^i &= \Gamma_{tor_{cos}} ([\mathbf{h}_a^L, \mathbf{h}_b^L, \mathbf{h}_c^L, \mathbf{h}_d^L, \mathbf{z}_i]) \forall (a,b,c,d) \in \mathbb{T}
    \\
    \sin \psi_{abcd}^i &= \Gamma_{tor_{sin}} ([\mathbf{h}_a^L, \mathbf{h}_b^L, \mathbf{h}_c^L, \mathbf{h}_d^L, \mathbf{z}_i]) \forall (a,b,c,d) \in \mathbb{T}
\end{align*}
Note that this encode-decoder architecture can be trained on arbitrary size of graphs and thus can be used to train a single model for multiple molecules.
\subsection{Dataset}
Our dataset is generated via a 100 ns molecular dynamics simulations of alanine dipeptide in water. The coordinates of the atoms of alanine dipeptide is extracted every 100 fs. For the propagator model (TFT), we perform some coarse-graining of the trajectories by taking average positions of the atoms over a window of $l_{cg}$ steps. This facilitates the model to learn the underlying dynamics instead of trying to predict the thermal vibrations present during MD at finite temperatures. For better performance in predicting sequencing longer than the input sequence length, we sample, we sample multiple long sequences starting from random points in the trajectory and extracting $l_{cg}(l_{in} + l_{seq}l_{out})$ frames each time where $l_{seq}$ is the length of the sequences we want to train on and $l_{in}$ and $l_{out}$ is the \textit{input chunk lenght} and \textit{output chunk length} of the TFT model. Each such sequence is a data point and we extract $train\_size + validation\_size$ number of such sequences. The idea to sample from random starting points is to make the model robust against variations in initial points. 
\subsection{Surrogate dynamics}
\subsubsection{Training}
The encoder-decoder network is trained first directly on randomly sampled frames from the MD trajectory. Mean square loss is used for training. We found the separate training to perform better than training both the models together. During the training of the propagator, the weights of the encoder-decoder model are frozen. The input to the propagator model is $\mathbf{z}_i$ as past covariates which comes from encoding the trajectory using the encoder. The loss functional is a sum of the error in the latent space prediction and the error in the final predicted topology. The latent space prediction uses gaussian regression with negative loss likelihood as the loss function. The error in the predicted topology is calculated by the mean squarred difference to the true values from MD, same as in the training the encoder-decoder network. Note that the different types of topological variables (bond, angles, torsions) are given different weight in the loss function.

\subsubsection{Prediction}
During prediction, $l_{in}$ number of steps are sampled from the MD trajectory as the starting point for the model. The model encoder encodes the model and while the TFT forecasts the future latent space points. Since we use gaussian likelihood regression as the loss function, the latent space points for the decoder must be sampled from a gaussian for which the model outputs the mean $\mu_{model}$ and variance $\sigma_{model}$. To replicate the MD dynamics at inference time, instead of $\sigma_{model}$, we use $\sigma_{md}$ which is variance directly computed from the latent space embedding of true MD trajectory (without coarse-graining). Additionally, to improve transition dynamics, we perform Independent Component Analysis (ICA) of a preliminary prediction run of $50000$ steps of the model and compare with the same analysis done on MD trajectory of same number of steps starting from a random point. The ratio between the mean of the eigenvalues from MD and model is used as a scaling factor for the variance $\sigma_{md}$ during prediction. Finally the sampled values are decoded via the decoder and topological values of interest are extracted.
\subsubsection{Results}
\begin{figure}[!htbp]
     \centering
     \begin{subfigure}[b]{0.48\textwidth}
         \centering
         \includegraphics[width=\textwidth]{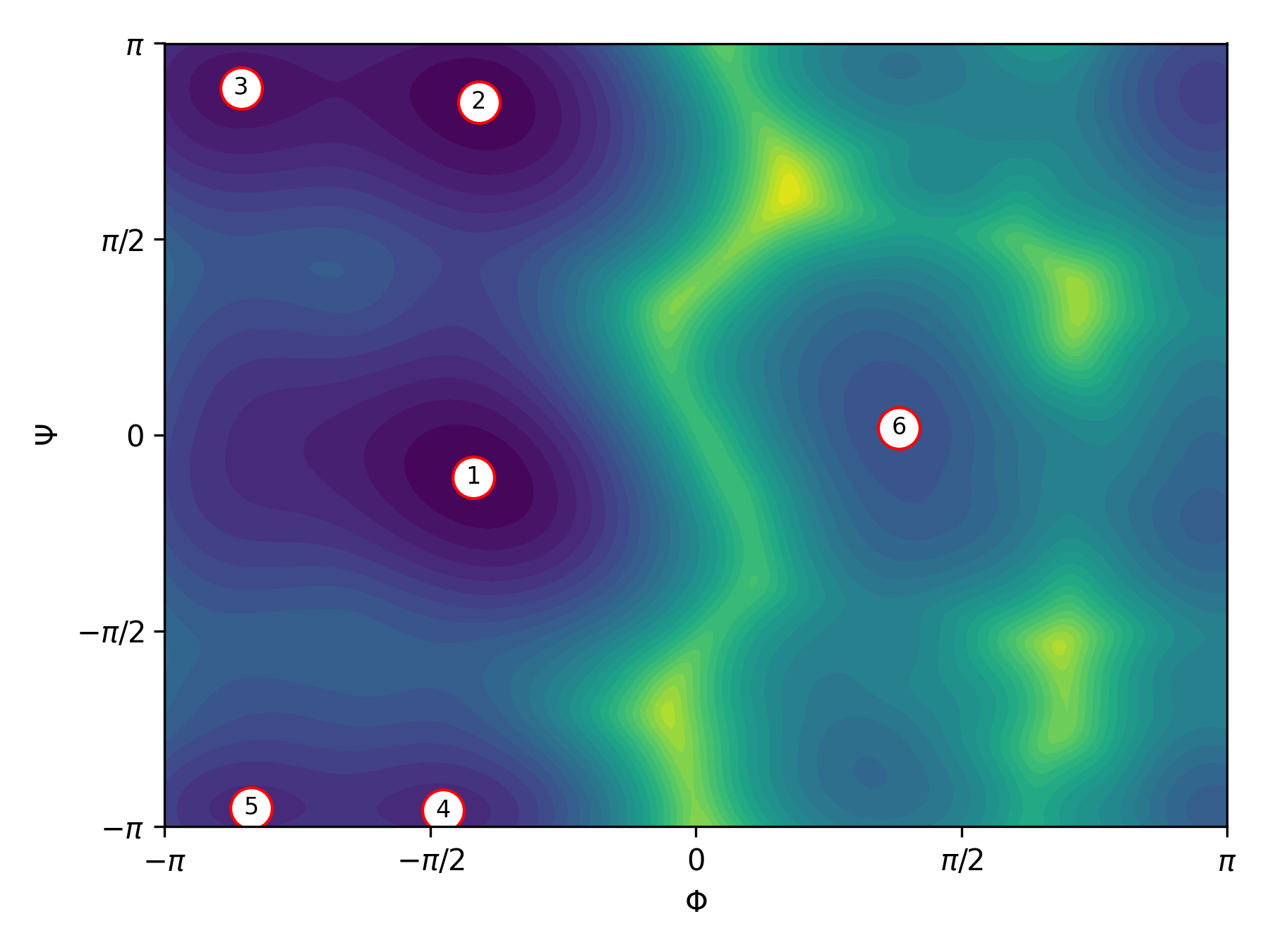}
         \caption{FES of true dynamics.}
         \label{fig:ala2_fes_true}
     \end{subfigure}
     \hfill
     \begin{subfigure}[b]{0.49\textwidth}
         \centering
         \includegraphics[width=\textwidth]{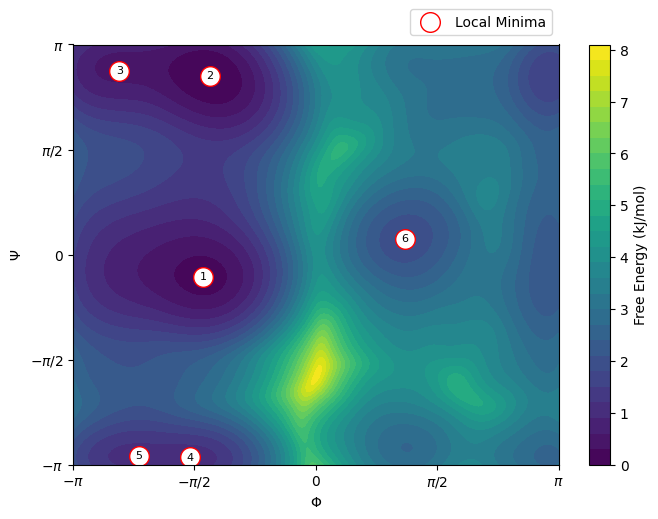}
         \caption{FES of predicted dynamics.}
         \label{fig:ala2_fes_pred}
     \end{subfigure}
     \caption{Comparison of Free Energy Surface (FES) of the dihedral angles $\phi$ and $\psi$ between true and surrogate dynamics.}
     \label{fig:ala2_fes}
\end{figure}

\begin{figure}[!htbp]
     \centering
     \begin{subfigure}[b]{0.49\textwidth}
         \centering
         \includegraphics[width=\textwidth]{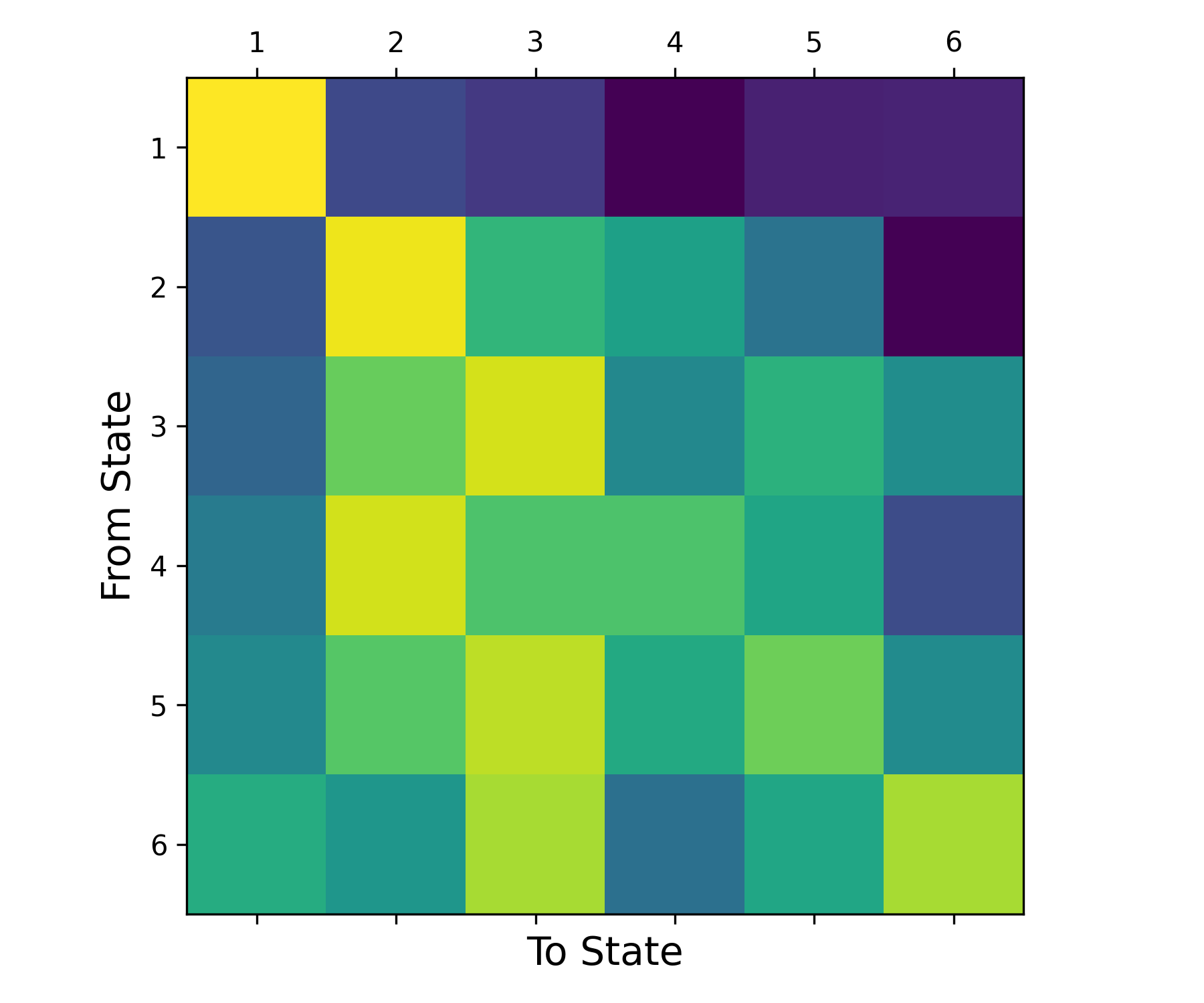}
         \caption{Transition probabilities of simulation dynamics.}
         \label{fig:ala2_tm_true}
     \end{subfigure}
     \hfill
     \begin{subfigure}[b]{0.49\textwidth}
         \centering
         \includegraphics[width=\textwidth]{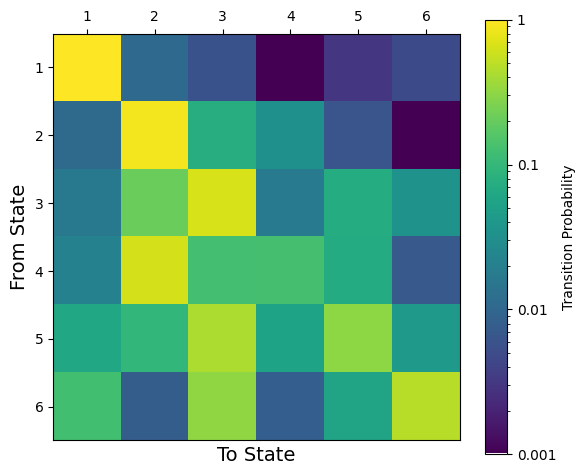}
         \caption{Transition probabilities of surrogate dynamics.}
         \label{fig:ala2_tm_pred}
     \end{subfigure}
     \caption{Comparison of Transition probabilities between simulation and surrogate dyanamics. States labeled in \cref{fig:ala2_fes}}
     \label{fig:ala2_tm}
\end{figure}

In \cref{fig:ala2_fes} we compare the Free Energy Surface spanned by the two dihedrals $\phi$ and $\psi$ which are the standard natural coordinates of the system. To obtain the FES, we perform kernel density approximation with a gaussian kernel of width 0.2 and bin size of 200. The surrogate model is in good overall agreement with the ground truth values. It is able to also capture the depth of the rate minimum at high $\phi$ values with decent accuracy.
\\
To look more closely at the accuracy of the transition statistics of the surrogate vs the simulated dynamics, we create a Markov State Model in the $\phi$-$\psi$ space. More specifically, to find the states, we start at 200 random points in the $\phi$-$\psi$ and run minimization algorithm using the BFGS method using the FES as the objective function. Then we cluster the final points using a proximity criterion with threshold of $0.15$ radians and label the top 6 populous clusters and the states (see minima labels in \cref{fig:ala2_fes}). We each point in the trajectory and assign it to the nearest state. Finally we use the number of counts the system transition from one state to other during the whole trajectory to build the transition counts matrix. We can then also calculate the transition probability matrix and the Mean First Passage Time (MFPT) matrix. The last one gives a measure number of step (400 fs for out setup) the simulation takes on average to observe a specific transition.
\\
We show the transition probabilities of the MD dynamics and surrogate model in \cref{fig:ala2_tm}. The surrogate dynamics capture the state transitions probabilities very closely hinting at the model learning a projection of true Boltzmann distribution of the state space of the dynamics.
\FloatBarrier
\section{Efficiency gain and Result physicality}
\begin{figure}[h!]
    \centering
    \includegraphics[width=0.9\linewidth]{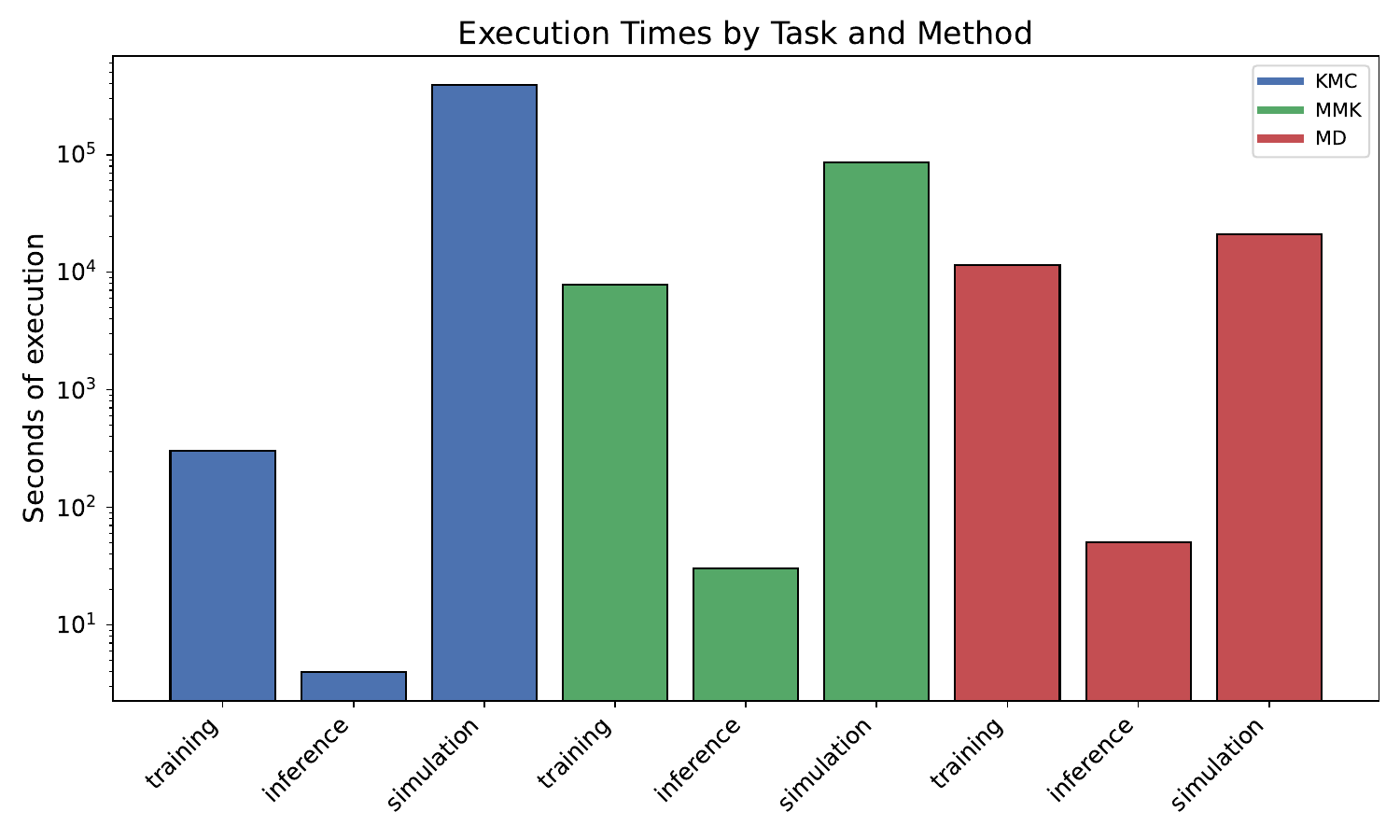}
    \caption{Time required for the execution of the simulation, the training of the model and the inference. }
    \label{fig:time}
\end{figure}

The previous chapters showcase the accuracy of the framework but do not shine sufficient light to the major advantage of our framework, namely its efficiency. Indicatively KMC simulations, would require from days to weeks to converge, while the TFT training required a fraction of that and its deployment an even more insignificant compute. Succintly, the comparative time requirements can be seen in \cref{fig:time} and for inference are without fail multiple orders of magnitude faster. 

\FloatBarrier
Further, it must be noted, that the first two simulations will eventually converge to an equilibrium change that is unchanged. It is of high importance that the surrogate model also reaches a steady unchangeable state. This is crucial as it proves that even if some accuracy loss occurs, the results do not deviate but instead converge, recreating physical stability as well as physical correctness.
\section{Outlook}
In conclusion, we have presented a versatile and robust framework, able to surrogate even complex and stochastic dynamics. Our time evolution model combines both long term predictions, captured by the transformer architecture, as well as local temporal relations, as determined by the LSTM. The model's ability to incorporate both time independent and time dependent covariates allows the inclusion of physical parameters, as well as time series that are less accurate but much faster to generate and still carry information about the target. Gated Units and Feature analysis naturally quantify the importance of the underlying physical parameters are seamlessly included.  The difficulty of the propagation task is alleviated by the use of advanced Autoencoder architectures. Additionally to robust and continuous dimensionality reduction, they also provide full configuration profile visualization which adds to the understanding of the investigated phenomenon.
The applicability of our framework was tested on multiple and diverse simulation methodologies and was found to perform well in all. Most impressively, we also measured a high computational efficiency increase, of at least three orders of magnitude, case dependent. 
We are therefore convinced that our framework can offer significant advantages in the simulation world by both accelerating dynamics investigation and preserving the intrinsic physical uncertainty of the simulations. 
\FloatBarrier
\section{Acknowledgements}
  I.K. acknowledges funding from the Project ProperPhotoMile, supported under the umbrella of SOLAR-ERA.NET Cofund 2 by The Spanish Ministry of Science and Education and the AEI under the project PCI2020-112185 and CDTI project number IDI-20210171; the Federal Ministry for Economic Affairs and Energy on the basis of a decision by the German Bundestag project number FKZ 03EE1070B and FKZ 03EE1070A and the Israel Ministry of Energy with project number 220-11-031. SOLAR-ERA.NET is supported by the European Commission within the EU Framework Programme for Research and Innovation HORIZON 2020 (Cofund ERA-NET Action, N° 786483).\\
  Further, M.G. acknowledges funding from the European Union's Horizon 2020 FETOPEN 2018–2020 program “LION-HEARTED” under grant agreement no. 828984.\\
  Finally, A.G. acknowledges financial support from TUM Innovation Network for Artificial Intelligence powered Multifunctional Material Design (ARTEMIS) and funding in the framework of Deutsche Forschungsgemeinschaft (DFG, German Research Foundation) under Germany's Excellence Strategy – EXC 2089/1 – 390776260 (e-conversion).
  
\clearpage
\bibliographystyle{unsrt}
\bibliography{references}

\end{document}